\documentclass[]{article}

\usepackage{times}

\usepackage{amsfonts}
\usepackage{amsmath}
\usepackage{bm}

\usepackage{graphicx}
\usepackage{subfigure}
\usepackage{multirow}

\usepackage{natbib}

\usepackage{hyperref}
\usepackage[accepted]{icml_like}
\usepackage{appendix}

\icmltitlerunning{Meta-learners' learning dynamics are unlike learners'}

\begin{document}

\twocolumn[
\icmltitle{Meta-learners' learning dynamics are unlike learners'}
\title{}

\vskip -0.6in
\author{
	\begin{tabular}[t]{c}
		{\bf Neil C. Rabinowitz\thanks{corresponding author}} 
		\\
		\\
		\today \\
		DeepMind \\
		\texttt{ncr@google.com} \\
	\end{tabular}
}

\date{}
\maketitle
\thispagestyle{empty}
]


\newcommand\todo[1]{\textcolor{red}{#1}}

\newcommand{\leg}{\bf}



\begin{abstract}
Meta-learning is a tool that allows us to build sample-efficient learning systems. Here we show that, once meta-trained, LSTM Meta-Learners aren't just \textit{faster} learners than their sample-inefficient deep learning (DL) and reinforcement learning (RL) brethren, but that they actually pursue fundamentally different \textit{learning trajectories}. We study their learning dynamics on three sets of structured tasks for which the corresponding learning dynamics of DL and RL systems have been previously described: linear regression \cite{saxe2013exact}, nonlinear regression \cite{rahaman2018spectral, xu2018training}, and contextual bandits \cite{schaul2019ray}. In each case, while sample-inefficient DL and RL Learners uncover the task structure in a staggered manner, meta-trained LSTM Meta-Learners uncover almost all task structure concurrently, congruent with the patterns expected from Bayes-optimal inference algorithms. This has implications for research areas wherever the learning behaviour itself is of interest, such as safety, curriculum design, and human-in-the-loop machine learning.
\end{abstract}

\section{Introduction}

It is widely recognized that the major paradigm of deep learning is sample inefficient. At the same time, many real-world tasks that an intelligent system needs to perform require learning at much faster time-scales, leveraging the data from only a few examples. Yet moving from sample-inefficient learning to sample-efficient learning may not be simply a matter of speeding things up.

Here we investigate one potential point of difference between sample-inefficient learners and sample-efficient ones: the \textit{order} in which they pick up structure in learning problems. 

We explore this through a series of studies on pairs of learning systems drawn from related but different paradigms: Deep Learning and Reinforcement Learning on the one hand, and (memory-based) meta-learning \cite{schmidhuber1996simple, thrun1998learning} on the other. The DL/RL networks and learning algorithms (``Learners'') provide us with instances of domain-general, sample-inefficient learning systems, while the meta-trained LSTMs provide us with instances of domain-specific, but sample-efficient learning systems. We refer to these LSTMs as ``Meta-Learners'', though we focus primarily on their inner loop of learning, after significant outer-loop training (i.e., after meta-learning) has occurred.

While previous effort has highlighted the sample efficiency, convergence, and generalisation properties of meta-learning systems \citep[e.g.][]{baxter1998theoretical, baxter2000model, hochreiter2001learning, vinyals2016matching, santoro2016meta, finn2017model, amit2018meta, grant2018recasting}, we explicitly factor these out. Instead, we consider whether Learners and Meta-Learners pursue the same \textit{learning trajectory}, i.e., whether they incrementally capture the same structure while learning the task.

To do this, we apply Learners and Meta-Learners to a series of three simple, structured tasks, within which the learning dynamics of Learners have been theoretically and empirically characterised \cite{saxe2013exact, rahaman2018spectral, xu2018training, schaul2019ray}. In each case, we measure whether the biases that Learners show during their learning process also manifest within Meta-Learners' learning process. In all three cases we find that they do not.

This work is relevant for several reasons. First, it gives us an indication of how different learning systems operate when they are not yet trained to convergence. While it is common practice to describe how learning systems behave once they have reached ``equilibrium'' (i.e., once training has converged), there are various reasons why we should anticipate ML systems will be deployed before reaching their theoretically-optimal performance. Most notably, real-world optimisation takes place under the constraints of finite resources, including finite compute and limited time. Moreover, we typically do not have the luxury of knowing exactly how far a given solution to a complex task is from the global optimum. A rich characterisation of how far learning systems progress on a resource budget gives us an indication of the kinds of errors that we should anticipate that they would make. Information we glean here could be valuable when evaluating the safety of deployed systems.

In addition to this, we add to the growing literature on how learning systems can build abstraction layers. In particular, previous work on meta-learning has shown how sample-inefficient ``outer optimisers'' can build sample-efficient ``inner optimisers'', and how the outer optimisers can draw from task information to bake novel priors into the inner ones \cite{rendell1987layered, thrun1998lifelong, giraud2004introduction, ravi2016optimization, grant2018recasting, amit2018meta, harrison2018meta}. Our contribution to this literature is to demonstrate an analogous phenomenon with the learning process itself: outer optimisers can configure inner optimisers to pursue \textit{learning trajectories} less available to a non-nested system.

Through this, our work puts into perspective previous research on the learning dynamics of neural network systems. It is tempting to interpret previous analyses \citep[e.g.\ ][]{saxe2013exact, arpit2017closer, rahaman2018spectral, xu2018training} as revealing something fundamental about neural networks' learning behaviour, for example, that neural networks will always learn the ``most important'', ``dominant'', or ``simple'' structure present in a task first. Our results show that these results are not as universal as they first seemed, as learning dynamics are also a function of the priors that a learner brings to a task.

The remainder of this document proceeds as follows. In Section~\ref{section:related-work}, we briefly review relevant work on learning dynamics, meta-learning, and the particular distinctions which have been made between Learners' and Meta-Learners' behaviours. We then walk through three experiments in turn. In Section~\ref{section:linear-regression}, following previous work of \citet{saxe2013exact}, we consider the learning dynamics of deep networks on linear regression tasks. In Section~\ref{section:nonlinear-regression}, following previous work of \citet{rahaman2018spectral} and \citet{xu2018training}, we consider the learning dynamics of deep networks on nonlinear regression tasks. Finally, in Section~\ref{section:contextual-bandits}, following previous work of \citet{schaul2019ray}, we consider the dynamics of reinforcement learning, specifically an interference phenomenon that occurs during on-policy learning of contextual policies. We conclude in Section~\ref{section:outer-loops} by characterising the outer learning dynamics by which the Meta-Learners are themselves configured.

\section{Related work}
\label{section:related-work}

There is a long history of interest in the learning dynamics of neural networks equipped with SGD and its variants (``Learners'', by our terminology). The field's principal goal in studying this has been to improve optimisation techniques. Many studies have thus investigated the learning process to extract convergence rates and guarantees for various optimisers, and well as to identify properties of optimisation landscapes for canonical tasks \cite{bottou2010large, dauphin2014identifying, su2014differential, choromanska2015loss, goodfellow2014qualitatively, schoenholz2016deep, wibisono2016variational, li2018visualizing, pennington2018emergence, yang2018physical, baity2018comparing}.

From this optimisation-centric perspective, intermediate stages of learning are generally viewed as obstacles to be overcome. As gradient descent methods are typically conceived as a local search in a hypothesis space, the point of learning is to find a better hypothesis than the current one. Thus one considers the gradient that takes one \textit{away} from the current parameters, how one can \textit{escape} from saddle points, and how to avoid \textit{getting stuck} in local minima or boring plateaux. Intermediate parameters (and their corresponding hypotheses) are rarely points of interest in and of themselves. This view is reflected in the standard procedure of plotting and presenting graphs of how loss decreases over the course of training, which reinforces the conception that the defining characteristic of an intermediate stage of learning is that it is not yet the end.

There have been a few notable areas of research where this conception has been openly challenged.

First, there have been efforts to characterise when \textit{particular} task substructures are learned during training. We take several of these as our starting point for investigation of sample-efficient learning dynamics, which are discussed throughout the text \cite{saxe2013exact, rahaman2018spectral, xu2018training, schaul2019ray}. We note three more cases from the supervised image classification literature. \citet{arpit2017closer} studied the order in which labels are learned by classifiers, finding that inputs with randomised labels are learned later than those with correct labels. \citet{achille2018critical} observed that certain structure could only be efficiently learned early in training, reminiscent of the phenomenon of critical periods in biological learning \cite{hensch2004critical}, while \citet{toneva2018empirical} found that the predicted labels for some inputs are regularly forgotten by many learners over the course of training, which the authors relate to the gradual construction of a maximum margin classifier \cite{soudry2018implicit}. There have likewise been several studies which have drawn attention to how various statistics change over the course of learning, such as representational geometry \citep[e.g.\ ][]{raghu2017expressive}, information geometry \citep[e.g.\ ][]{shwartz2017opening, achille2018emergence, chaudhari2018stochastic, saxe2018information}, and generalisation \citep[e.g.\ ][]{prechelt1998early, advani2017high, lampinen2018analytic}.

Second, learning dynamics are a central focus in the field of curriculum learning \cite{elman1993learning, bengio2009curriculum, weinshall2018theory}, where it is necessary to identify what a learner has and has not yet grasped in order to select data for the next stage for training.

Finally, the intermediate behaviour of learners plays a crucial role in reinforcement learning (as well as the more general field of active learning \cite{lewis1994sequential, cohn1996active, settles2012active}). Here there exists a feedback loop between the current stage of learning (in RL, via the policy) and the data distribution being trained on. An agent's behaviour during intermediate stages of learning is directly responsible for generating the opportunities to change the policy. This is a core reason for the critical research agenda on deep exploration \citep[e.g.\ ][]{jaksch2010near, osband2016deep, houthooft2016vime, pathak2017curiosity, tang2017exploration, fortunato2017noisy}.

Aside from related work on learning dynamics, this paper connects to a literature that is beginning to probe the behaviour of meta-learners. There now exists a range of novel meta-learning systems that exploit memory-based networks \cite{hochreiter2001learning, santoro2016meta, wang2016learning, duan2016rl2, mishra2017metalearning}---such as the LSTM Meta-Learners we consider here---as well as gradient-based inner optimisation \cite{finn2017model, rusu2018meta}, evolution \cite{hinton1987learning, fernando2018meta}, and learned optimisers \cite{andrychowicz2016learning, chen2017learning, li2016learning, ravi2016optimization}, with a range of applications that are too numerous to survey here. To the best of our knowledge, however, comparatively less work has been done to explore the detailed manner in which the meta-learned learners progressively acquire mastery over task structure during an episode. Several papers have identified relationships between meta-learners' inner learning processes and Bayes-optimal algorithms, either empirically \cite{kanitscheider2017training, chen2017learning, rabinowitz2018machine}, or theoretically \cite{baxter1998theoretical, baxter2000model, grant2018recasting}. Some work has been done to explore how the behaviour of these systems changes over the course of inner learning in specific domains: \citet{wang2016learning} identified the adoption of different effective learning rates in meta-reinforcement learning; \citet{eslami2018neural}, \citet{rabinowitz2018machine}, and \citet{garnelo2018neural} characterised how inner learners' uncertainty decreases over inner learning in networks trained on conditional scene generation, intuitive psychology, and GP-like prediction respectively; while \citet{kanitscheider2017training} and \citet{gupta2018meta} characterised inner exploration policies (and the limits thereof \cite{dhiman2018critical}). Our work adds to this literature by empirically characterising inner learners' learning process on simple problems with clear known structure, and differentiating these dynamics from those of DL and RL Learners.

Finally, a goal of this work is to provide clear examples of how \textit{sample-inefficient} learning may, at the behavioural level, look radically different from \textit{sample-efficient} learning. This links directly to the question of how we should relate machine learning to human learning. The manner in which humans learn has long served as core inspiration for building machine learning systems \cite{turing1950computing, rosenblatt1958perceptron, schank1972conceptual, fukushima1982neocognitron}, and remains an aspirational standard for much of AI research \citep[e.g.][]{mnih2015human, lake2015human, vinyals2016matching, lake2017building}. This, of course, applies to many of the core learning paradigms, whose origins can be traced in part to settings in which humans learn, such as supervised learning \cite{rosenblatt1958perceptron}, unsupervised learning \cite{hebb1949organization, barlow1989unsupervised, hinton1986learning}, reinforcement learning \cite{thorndike1911animal, sutton1998reinforcement}, imitation learning \cite{piaget1952play, davis1973imitation, galef1988imitation, hayes1994robot, schaal2003computational}, and curriculum learning \cite{elman1993learning}. But there has also long been a desire to connect human and machine learning at a finer grain. Early connectionist models proposed relationships between the dynamics of human learning during development and the dynamics of artificial neural networks during training \cite{plunkett1992connectionism, mcclelland1995connectionist, rogers2004semantic, saxe2018mathematical}. Notwithstanding prevailing discussions about whether the mechanisms of backpropagation are biologically plausible \cite{crick1989recent, stork1989backpropagation, lillicrap2016random, marblestone2016toward,  guerguiev2017towards}, the idea that humans and machines may show similar learning dynamics at a \textit{behavioural} level remain current, with parallels recently being proposed for perceptual learning \cite{achille2018critical} and concept acquisition \cite{saxe2018mathematical}. Moreover, a number of similarities have been identified between the \textit{representations} produced by machine and human learning \citep[e.g.][]{olshausen1996emergence, yamins2014performance, khaligh2014deep, gucclu2015deep, nayebi2018task, yang2018physical}. This line of research is suggestive that there may exist a level of abstraction within which the dynamical processes of deep learning and human learning align.

Despite this argument, there remain huge points of difference between how we and current deep learning systems learn to solve many of the tasks we face. \citet{lake2017building} constructs a sweeping catalogue of these, highlighting, most relevantly here, humans' sample-efficiency, as well as features of the solutions we produce such as their generalisability, deployment of causality, and compositionality. As the ML community increasingly innovates on techniques that yield more sample-efficient learning in specialised domains, our contribution to this field lies in raising the hypothesis: when humans and machines learn with radically different sample efficiencies, their \textit{learning trajectories} may well end up being very different.

\section{Experiment 1: Linear regression}
\label{section:linear-regression}

In our first experiment, we study a phenomenon originally described by \citet{saxe2013exact}, and extended in a number of subsequent studies \cite{advani2017high, saxe2018mathematical, stock2018learning,
bernacchia2018exact, lampinen2018analytic}. These results begin with an analytic demonstration that when one trains deep linear networks on linear regression problems using stochastic gradient descent (SGD), they learn the target function with a stereotyped set of dynamics.

More precisely, we define the target regression problem to be of the form $\bm{y} = \bm{W} \bm{x} + \bm{\varepsilon}$, where $\bm{\varepsilon} \sim \mathcal{N}(\bm{0}, \bm{\Sigma_{\varepsilon}})$. A deep linear network is a feedforward network without any nonlinearities, parameterised by a set of weight matrices $\bm{W_i}$, and whose outputs are of the form $\hat{\bm{y}} = \left( \prod_{i=1}^n \bm{W_i} \right) \bm{x}$. Given this setup, \citet{saxe2013exact} showed that over the course of training, the network will learn the singular modes of $\bm{W}$ in order of their singular values. Thus, if we define the singular value decomposition (SVD) of the target matrix via $\bm{W} = \bm{U} \bm{S} \bm{V}^\top = \sum_i s^{(i)} \bm{u^{(i)}} \bm{v}^{\bm{(i)}\top}$, then the time course of learning of each singular mode, $\bm{u^{(i)}} \bm{v}^{\bm{(i)}^\top}$, is sigmoidal, with a time constant inversely proportional to the singular value $s^{(i)}$. In both this, and subsequent work, it was shown empirically that these results extend to the training of deep nonlinear networks using stochastic gradient descent.

In what follows, we first replicate empirical results for an SGD-based deep Learner for completeness (Section~\ref{subsection:linear-regression:learner}), setting up the standard experimental parameters and protocols. We then consider the dynamics by which a sample-efficient Meta-Learner---trained to learn linear regression problems---acquires structure while learning problems of the same form (Section~\ref{subsection:linear-regression:meta-learner}). We save analysis of the learning process by which the Meta-Learner itself is constructed to Section~\ref{section:outer-loops}.

\subsection{Learner}
\label{subsection:linear-regression:learner}

\subsubsection{Setup}

We consider linear regression problems of the form $\bm{y} = \bm{W} \bm{x} + \bm{\varepsilon}$, where $\bm{x} \in \mathbb{R}^{N_x}$ and $\bm{y}, \bm{\varepsilon} \in \mathbb{R}^{N_y}$. We run experiments both where $N_x = N_y = 2$ and $N_x = N_y = 5$ (i.e., 2D and 5D respectively). In all cases, we use isotropic noise, with $\bm{\varepsilon} \sim \mathcal{N}(\bm{0}, 0.01^2 \bm{I})$.

We train Learners on a single task, defined by sampling a random ground-truth weight matrix $\bm{W}$ with a fixed spectrum. To do this, we sample a matrix from the zero-mean, isotropic matrix-normal distribution $\mathcal{MN}(\bm{0}, \bm{I}, \bm{I})$, compute its SVD, then replace the spectrum with the desired one. This weight matrix is then held fixed for the duration of training. All data is procedurally-generated (i.e., there is no fixed training set), with inputs $\bm{x} \sim \mathcal{N}(\bm{0}, \bm{I})$.

For our Learners, we use 2-layer linear MLPs equipped with SGD, the details of which can be found in Appendix~\ref{appendix:architectures:linear:learner}. The objective, as standard, is to minimise the $l_2$ loss between the network's prediction, $\bm{\hat{y}}$, and the target, $\bm{y}$. As per previous work \cite{saxe2013exact, advani2017high, lampinen2018analytic}, there are a wide range of deep network architectures and initialisations under which the same results hold, and we find our results similarly robust.

\subsubsection{How to assess learning dynamics}
\label{subsubsection:linear-regression:learner:how}

Since we are interested in the learning dynamics of the network, we use the subscript $t$ to denote the training step. Writing the output of the ($L = 2$ layer) network at step $t$ as:
\begin{align}
\bm{\hat{y}_t} &= \left( \prod_{l=1}^{L} \bm{W_{l,t}} \right) \bm{x} \\
				&= \bm{\hat{W}_t} \bm{x}
\end{align}
we obtain the effective linear function executed by the network at this training step as $\bm{\hat{W}_t}$. We then project the effective linear weight matrix, $\bm{\hat{W}_t}$, onto the coordinate system induced by the SVD of the target weight matrix, $\bm{W}$. If we write the decomposition as $\bm{W} = \bm{U} \bm{S} \bm{V}^\top$, then the spectrum of $\bm{W}$ can be expressed as $\bm{s} = \mathrm{diag}(\bm{U}^\top \bm{W} \bm{V})$. Thus, to determine the effective progress at training step $t$, we compute the network's \textit{effective spectrum}:
\begin{equation}
	\label{eq:linear:learner:effective-spectra}
	\bm{\hat{s}}_t = \mathrm{diag} \left( \bm{U}^\top \bm{\hat{W}_t} \bm{V} \right)
\end{equation}

We ignore the non-diagonal terms, which carry residuals not aligned to the target task's singular modes. In what follows, we primarily consider the learning dynamics of individual singular modes, $\hat{s}_{k,t}$ (where $k$ indexes the mode), rather than the learning dynamics of joint spectra.

\subsubsection{Results}

\begin{figure*}[th]
\begin{centering}
	\includegraphics[width=18cm]{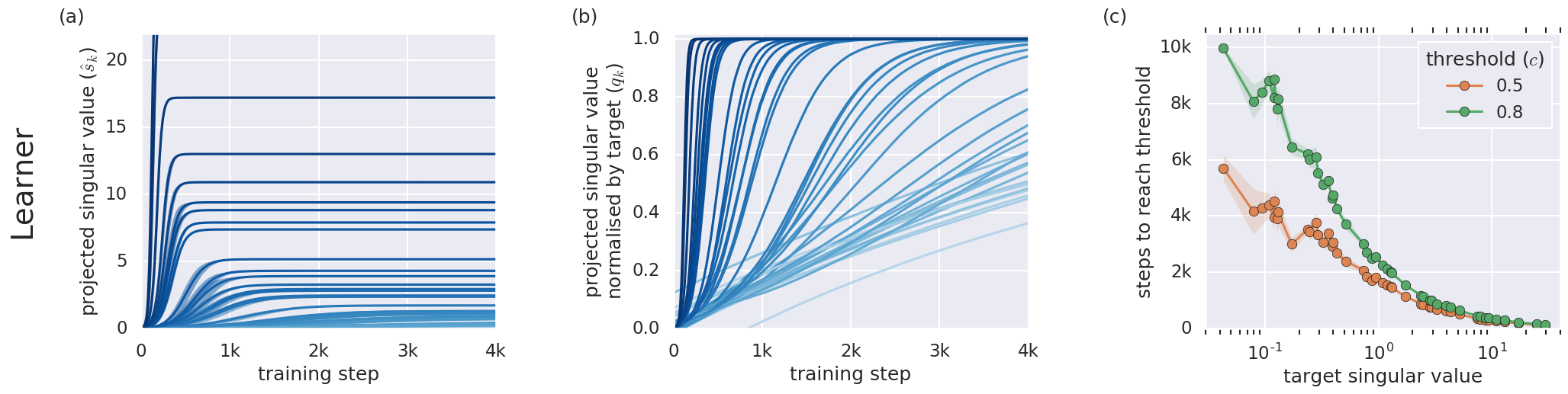}
	\caption{
		\textbf{(a)} Learning dynamics of the Learner ($\hat{s}_{k,t}$) for singular modes of different singular values. Modes with larger singular values are darker, and converge faster, as per \citet{saxe2013exact}. Data are accrued from a number of Learners trained on different 2D regression problems with a range of target spectra.
		\textbf{(b)} As in (a), but showing proportions of the target value, $q_k(t) = \hat{s}_{k,t} / s_{k}$. This makes the difference in learning speed more visible.
		\textbf{(c)} Training steps required for the Learner to reach a threshold of performance where $q_k(t) > c$, for different cutoff values $c$. This demonstrates that the learning speed of singular modes is slower for smaller singular values. Shaded areas here, and in the rest of the paper, denote standard errors across networks.
	\label{fig:linear:learner:individual-modes}
	}
\par\end{centering}
\end{figure*}

Fig~\ref{fig:linear:learner:individual-modes}a shows the Learner's learning dynamics for individual singular modes. As per \citet{saxe2013exact}, singular modes of $\bm{W}$ with large singular values (darker colours, larger asymptotic values) are learned faster, i.e.\ with sigmoidal learning curves that converge quicker. Conversely, singular modes with smaller singular values (lighter colours, smaller asymptotic values) are learned slower. These dynamics are even more visible when we primarily consider the learned singular values as a proportion of the target singular value, $q_k(t) = \hat{s}_{k,t} / s_{k}$, in Fig~\ref{fig:linear:learner:individual-modes}b. Finally, we quantify the learning rate of these singular modes by measuring the number of training steps it takes for the learned singular modes to reach a threshold proportion of the target value, $q_k(t) > c$, in Fig~\ref{fig:linear:learner:individual-modes}c. Here it can be seen that the larger the value of the singular mode to be estimated, the fewer steps it takes.

Similar results are shown for the 5D case in Appendix Fig~\ref{appendix:fig:linear:5D}a, and for nonlinear deep nets in Appendix Fig~\ref{appendix:fig:linear:nonlinear-net}.

\subsection{Meta-Learner}
\label{subsection:linear-regression:meta-learner}

\subsubsection{Setup}
\label{subsubsection:linear-regression:meta-learner:setup}

We sought to compare these learning dynamics with those of a sample-efficient meta-learner, which had been explicitly configured to be able to learn new linear regression problems. 

To do this, we (``outer''-)trained an LSTM Meta-Learner (architecture details in Appendix~\ref{appendix:architectures:linear:meta-learner}), using the same rough procedure as \citet{hochreiter2001learning} and \citet{santoro2016meta}, as follows.

On each length-$T$ episode $i$ during outer-training, we presented the Meta-Learner with a \textit{new} regression problem, by sampling a new target weight matrix $\bm{W^{(i)}}$. We then sampled a sequence of inputs, $\lbrace \bm{x^{(i)}_t} \rbrace_{t=1}^{T}$, where each $\bm{x^{(i)}_t} \sim \mathcal{N}(\bm{0}, \bm{I})$ for each time step $t \in \lbrace 1, ..., T \rbrace$. In turn, we computed the corresponding targets, $\bm{y^{(i)}_t} = \bm{W^{(i)}} \bm{x^{(i)}_t} + \bm{\varepsilon^{(i)}_t}$, with each $\bm{\varepsilon^{(i)}_t} \sim \mathcal{N}(0, 0.01^2 \bm{I})$.

At each time point $t$ in the episode, the objective of the LSTM Meta-Learner is to predict the target $\bm{y^{(i)}_t}$, given $\bm{x^{(i)}_t}$ and the history of previous inputs and targets, $\lbrace \left( \bm{x^{(i)}_{t^\prime}}, \bm{y^{(i)}_{t^\prime}} \right) \rbrace_{t^\prime = 1}^{t-1}$. We implement this by feeding into the LSTM at time $t$ a concatenation of $\bm{x^{(i)}_t}$ and $\bm{y^{(i)}_{t-1}}$ (setting $\bm{y^{(i)}_0} = \bm{0}$), and linearly reading out a prediction $\bm{\hat{y}^{(i)}_t}$ from the LSTM at the same time step. 

To outer-train the LSTM itself, we minimised the average of the $l_2$ losses over the course of the episode, i.e., the quantity:
\begin{equation}
				\mathcal{L}^{(i)} = \frac{1}{T} \sum_{t=1}^T ||\bm{\hat{y}^{(i)}_t} - \bm{y^{(i)}_t}||^2
\end{equation}
using BPTT and episode lengths of $T=20$. During this outer-training process, we sampled target weight matrices from a fixed distribution. Unless stated otherwise, we used a standard matrix-normal distribution, with $\bm{W^{(i)}} \sim \mathcal{MN}(\bm{0}, \bm{I}, \bm{I})$. For consistency between the outer-learning algorithm and the Learner described in the previous Section, we trained the Meta-Learner with SGD.

\subsubsection{Inner and outer learning dynamics}

We note that the Meta-Learner has \textit{two} learning dynamics: the \textit{inner dynamics} by which the configured LSTM learns a solution to a new regression problem; and the \textit{outer dynamics} by which the LSTM itself is shaped in order to be able to do regression at all. While the inner learning dynamics are sample efficient and occur through changes in the LSTM's activations, the outer learning dynamics are sample inefficient and occur through changes in the LSTM's weights.

Our primary focus in this manuscript is a comparison between how two learning systems approach single regression problems. As such, we concentrate here on the inner dynamics of the Meta-Learner, once outer-training has produced a sufficiently performant system. This allows us to make direct comparison between the process by which sample-inefficient Learners and sample-efficient Meta-Learners come to discover and capitalise on structure in the same learning task. We leave all analysis of outer-loop behaviour to Section~\ref{section:outer-loops}.

\subsubsection{How to assess learning dynamics}
\label{subsubsection:linear-regression:meta-learner:how}

We thus consider here a Meta-Learner LSTM with fixed weights. Our goal is to estimate the effective function realised by the LSTM at iteration $t$ of an episode $i$, which we denote $\hat{g}^{(i)}_t: \mathbb{R}^{N_x} \rightarrow \mathbb{R}^{N_y}$, and determine how this function changes with more observations (i.e., as $t$ increases).

We begin by sampling a target weight matrix for the episode, $\bm{W^{(i)}}$, with a spectrum of interest. For each time point, $t$, we sample a \textit{fixed} sequence of prior observations, $\lbrace \left( \bm{x^{(i)}_{t^\prime}}, \bm{y^{(i)}_{t^\prime}} \right) \rbrace_{t^\prime = 1}^{t-1}$. Using this sequence to determine the LSTM's hidden state, we then compute a linear approximation of $\hat{g}^{(i)}_t(\cdot)$ as follows: we sample a probe set of $N$ different values of $\bm{x^{(i)}_t}$ from $\mathcal{N}(\bm{0}, \bm{I})$; compute the respective values of $\bm{\hat{y}^{(i)}_t}$; and use linear least squares regression on this probe set to approximate $\bm{\hat{y}^{(i)}_t} \approx \bm{\hat{W}^{(i)}_t} \bm{x^{(i)}_t}$. 

We focus our attention on how the estimates $\bm{\hat{W}^{(i)}_t}$ project onto the singular modes of the episode ground-truth matrix $\bm{W^{(i)}} = \bm{U^{(i)}} \bm{S^{(i)}} \bm{V}^{\bm{(i)}\top}$, via the effective spectra:
\begin{equation}
\bm{\hat{s}^{(i)}_t} = \mathrm{diag} \left( \bm{U}^{\bm{(i)}\top} \bm{\hat{W}^{(i)}_t} \bm{V^{(i)}} \right)
\end{equation}

By tracking the effective spectra over the course of (inner) learning, we obtain an analogous measurement to that computed for the Learner in Equation~\eqref{eq:linear:learner:effective-spectra}.

For each time, $t$, and episode $i$, we use $N=100$ samples to estimate $\bm{\hat{W}^{(i)}_t}$, all with the same fixed sequence of prior observations. When studying the learning dynamics for particular target spectra, we sample new target matrices $\bm{W^{(i)}}$ with the same spectra, and for each, compute the inner dynamics over which the singular values are learned (using new input samples as well). For each time, $t$, we average the effective spectra, $\bm{\hat{s}^{(i)}_t}$ over 100 episodes (where each episode has the same target spectrum, $\bm{s^{(i)}}$, but uniquely-rotated target matrices and unique input sequences) and report statistics over 10 independently-trained instances of Meta-Learner LSTMs.

\subsubsection{Results: in-distribution}

\begin{figure*}[th]
\begin{centering}
	\includegraphics[width=13cm]{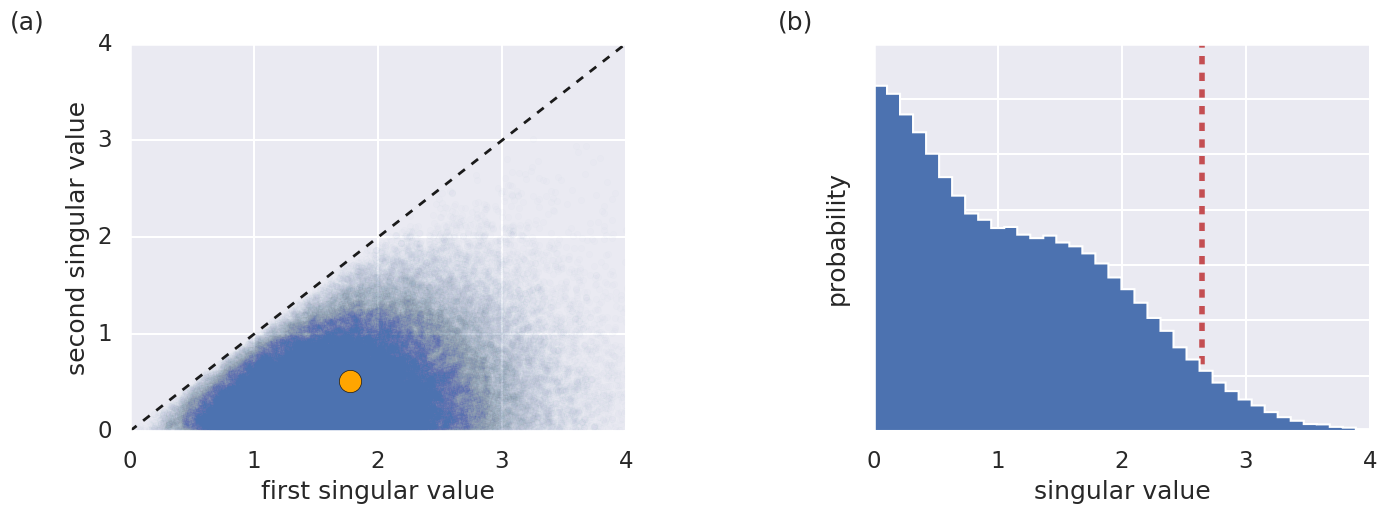}
	\caption{
		\textbf{(a)} Samples from the joint distribution over first and second singular values for $2 \times 2$ matrices $\bm{W^{(i)}} \sim \mathcal{MN}(\bm{0}, \bm{I}, \bm{I})$. Orange dot shows mean spectrum.
		\textbf{(b)} Marginal distribution of all singular values. Red vertical line shows the $95^{\mathrm{th}}$ percentile.
	\label{fig:linear:distribution-of-spectra}
				}
\par\end{centering}
\end{figure*}

The Meta-Learner is outer-trained on weight matrices drawn from the standard matrix normal distribution. This induces a particular distribution over singular values, shown in Fig~\ref{fig:linear:distribution-of-spectra}. We concentrate first on the Meta-Learner's learning behaviour for singular values up to the 95th percentile of this distribution. Learning on more extreme singular values (which we consider out-of-distribution) is characterised in Section~\ref{section:linear-regression:meta-learner:ood} below.

\begin{figure*}[th]
\begin{centering}
	\includegraphics[width=18cm]{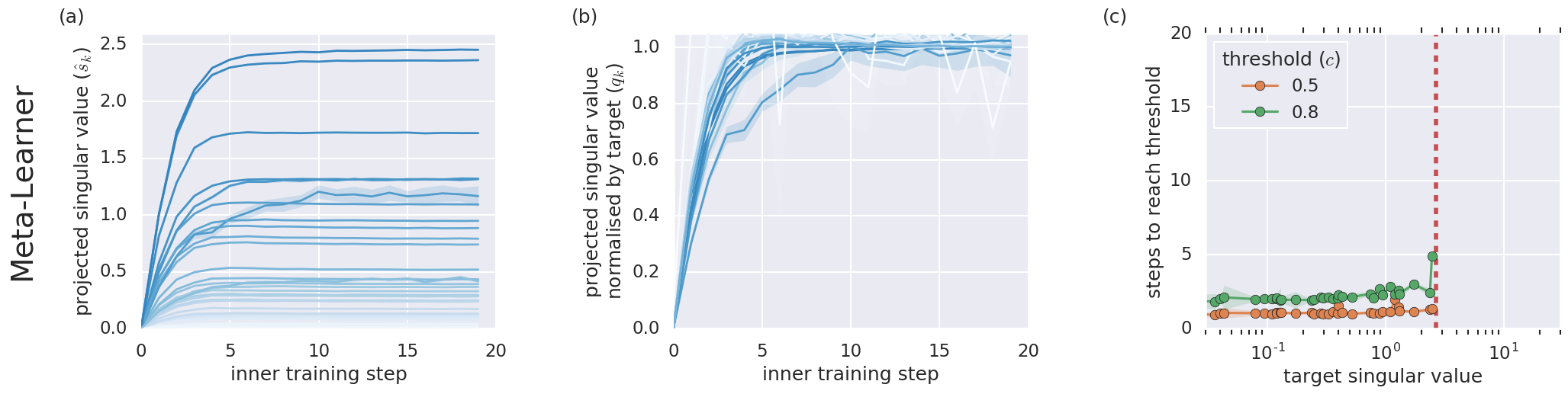}
	\caption{
    	Learning dynamics of the Meta-Learner for singular modes of different singular values, as in Fig~\ref{fig:linear:learner:individual-modes}.
	\label{fig:linear:meta-learner:individual-modes}
	}
\par\end{centering}
\end{figure*}

Fig~\ref{fig:linear:meta-learner:individual-modes} shows the Meta-Learner's learning dynamics for individual singular modes, in the same form as Fig~\ref{fig:linear:learner:individual-modes}. We note two major differences. First, as expected, the Meta-Learner is dramatically more sample efficient than the Learner (by a factor of $\mathcal{O}(1000)$). However, when we factor this out, we see a second important difference: the Meta-Learner does \textit{not} show the Learner's pattern of learning singular modes with large singular values faster. Rather, the Meta-Learner learns all in-distribution singular modes at roughly the same rate.

Putting these changes into perspective, a $50\times$ decrease in a mode's singular value causes the Learner to take roughly $9\times$ as long to learn it, while it causes the Meta-Learner to take only $1.05\times$ as long.

We show similar results for 5D matrices ($N_x = N_y = 5$) in the appendix, as Figs~\ref{appendix:fig:linear:5D}b-c.

\begin{figure*}[th]
\begin{centering}
	\includegraphics[width=13cm]{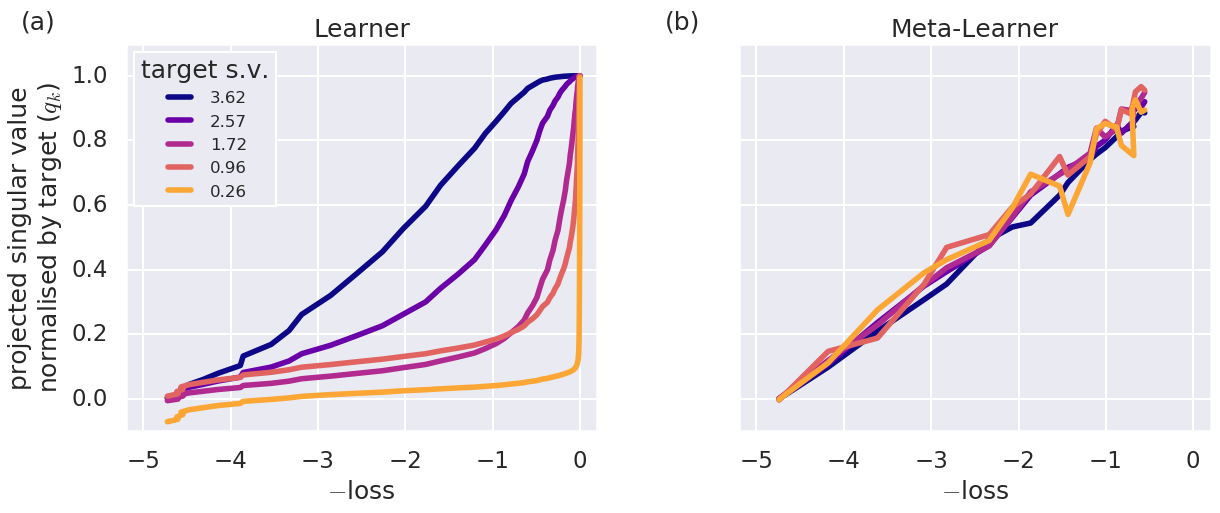}
	\caption{
		Effective spectrum of a 5D linear regression problem over the course of training, for \textbf{(a)} the Learner, and \textbf{(b)} the Meta-Learner. The x-axes show training progress (shown as the negative of the loss), which improves as one moves rightwards. Colours show the learning progress of each singular mode, as the proportion of the target value, $q_k(t)$. Results are shown averaged over 100 target matrices $\bm{W^{(i)}}$, all of which have the same spectrum (which we selected as the expected spectrum of matrices drawn from $\mathcal{MN}(\bm{0}, \bm{I}, \bm{I})$).
	\label{fig:linear:spectrum-vs-performance}
	}
\par\end{centering}
\end{figure*}

An instructive way to view the difference between the Learner's behaviour and the Meta-Learner's is through the evolution of the effective spectrum over the course of training. In this way, one can consider Learners and Meta-Learners at equal levels of performance, and compare what components of the task they have captured at this stage of learning. We show this in Fig~\ref{fig:linear:spectrum-vs-performance} for a 5D linear regression problem. Here it can be clearly seen how the Learner acquires the dominant singular structure first, while the Meta-Learner acquires all the singular structure concurrently.

\subsubsection{Results: out-of-distribution}
\label{section:linear-regression:meta-learner:ood}

\begin{figure*}[th]
\begin{centering}
	\includegraphics[width=18cm]{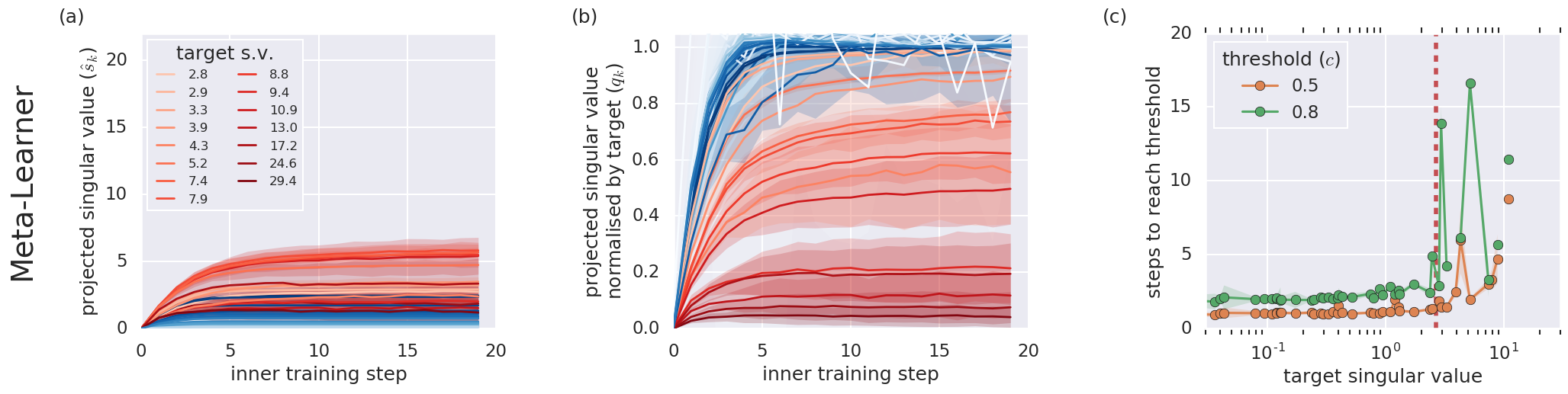}
	\caption{
		Learning dynamics of the Meta-Learner for singular modes of different singular values, as in Fig~\ref{fig:linear:meta-learner:individual-modes}. Blue lines in (a) and (b) are the dynamics of learning for within-distribution singular values, as in Fig~\ref{fig:linear:meta-learner:individual-modes}; red lines show learning of out-of-distribution singular values. Red vertical line in (c) shows the $95^{\mathrm{th}}$ percentile of the distribution of singular values, as in Fig~\ref{fig:linear:distribution-of-spectra}b.
		\label{fig:linear:meta-learner:individual-modes-ood}
		}
\par\end{centering}
\end{figure*}

As mentioned above, the Meta-Learner is outer-trained on matrices with a particular distribution of singular values. When we step outside this---by presenting the Meta-Learner with a target matrix with singular modes greater than the $95^\mathrm{th}$ percentile---the Meta-Learner progressively fails to learn the full extent of the singular mode (Fig~\ref{fig:linear:meta-learner:individual-modes-ood}). 

One potential cause of this behaviour is saturation of the LSTM outside of its natural operating regime when the target outputs are too large. However, this regime is an (outer) learned property: when we outer-train the Meta-Learner on weight matrices from a scaled matrix normal distribution, with $\bm{W^{(i)}}\sim \mathcal{MN}(\bm{0}, \alpha^2 \bm{I}, \alpha^2 \bm{I})$, the patterns seen in Fig~\ref{fig:linear:meta-learner:individual-modes-ood} shift approximately to match the scale of singular modes within the distribution (e.g.\  Fig~\ref{appendix:fig:linear:meta-learner:different-scales}).

Congruent with this hypothesis, the Meta-Learner was generally competent at learning singular values \textit{smaller} than a lower cutoff of the outer-training distribution. When we outer-trained the Meta-Learner on more exotic distributions of matrices---such as enforcing the distribution of singular values to be uniform over a range $[s_{\mathrm{min}}, s_{\mathrm{max}}]$---the Meta-Learner was equally fast (and competent) at learning singular values smaller than $s_{\mathrm{min}}$ as those within $[s_{\mathrm{min}}, s_{\mathrm{max}}]$.

Finally, one might ask whether the difference in learning dynamics between the Learner and Meta-Learner are due to different parameterisations: while the Learner uses a feedforward network, the Meta-Learner uses an LSTM. To test this hypothesis, we trained a Learner using the same LSTM architecture as the Meta-Learner (albeit solving a single linear regression problem only, i.e.\ one where the target matrix is always the same across episodes). This did not qualitatively affect the learning dynamics of the Learner (Fig~\ref{appendix:fig:linear:learner:lstm:individual_modes}).

\subsubsection{Relationship to Bayes-optimal inference}
\label{subsubsection:linear-regression:bayes-optimal}

A number of links have been drawn between meta-learning and Bayes-optimal inference, in particular, that over outer-training, the inner loop should come to approximate amortised Bayes-optimal inference \cite{baxter1998theoretical, baxter2000model, grant2018recasting}. Indeed, we show here that Bayes-optimal inference follows the same qualitative learning dynamics as the Meta-Learner on solving new (in-distribution) linear regression problems.

We assume we have a Bayes-optimal observer performing multivariate inference, with the correct prior over $\bm{W}$, $P(\bm{W}) = \mathcal{MN}(\bm{0}, \bm{I}, \bm{I})$, and the correct ground-truth noise covariance, $\bm{\Sigma_\varepsilon} = \sigma_\varepsilon^2 \bm{I}$. Given this, the posterior over $\bm{W}$ after $t$ observations takes the form \cite{bishop2006pattern}:
\begin{align}
	P(\bm{W} \vert \bm{Y_t}, \bm{X_t}) \;&=\;
	\mathcal{MN}(\bm{\bar{W}_t},\, \bm{\Lambda_t}^{-1},\, \sigma_e^2 \bm{I}) 
	\label{eq:linear:bayes:1}
	\\
	\bm{\bar{W}_t} \;&=\; 
	(\bm{X_t}^\top \bm{X_t} + \sigma_e^2 \bm{I})^{-1} \bm{X_t}^\top \bm{Y_t} \\
	\bm{\Lambda_t} \;&=\;
	\bm{X_t}^\top \bm{X_t} + \sigma_e^2 \bm{I}
	\label{eq:linear:bayes:3}
\end{align}
where $\bm{X_t}$ and $\bm{Y_t}$ are matrices of $t$ observed inputs and outputs respectively.

\begin{figure*}[th]
\begin{centering}
  \includegraphics[width=18cm]{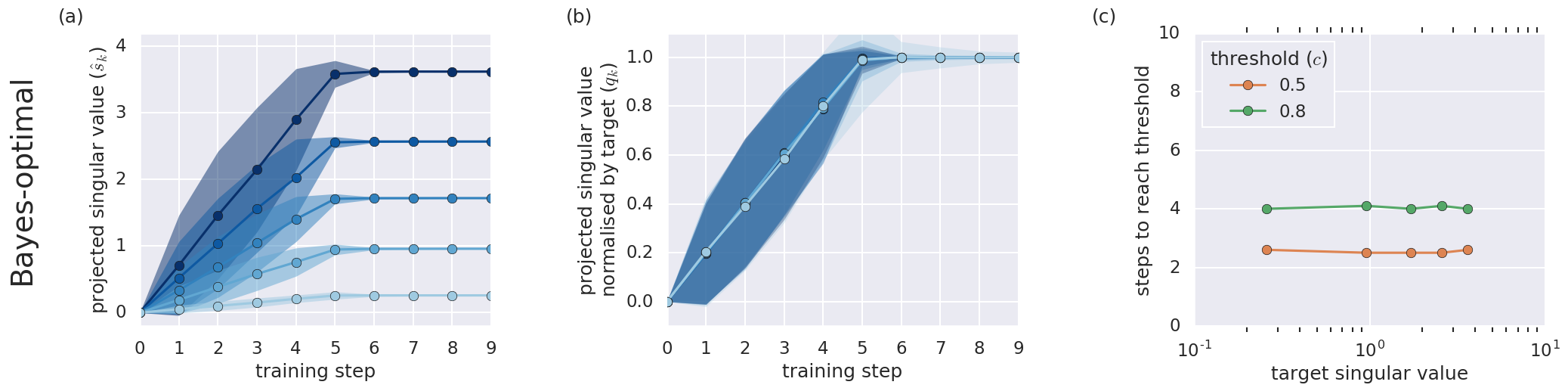}
	\caption{
	  Learning dynamics of Bayesian multi-variate linear regression for a 5D problem (with spectrum as in Fig~\ref{fig:linear:spectrum-vs-performance}). Results shown as for Fig~\ref{fig:linear:meta-learner:individual-modes}.
		\label{fig:linear:optimal}
		}
\par\end{centering}
\end{figure*}

In Fig~\ref{fig:linear:optimal}, we show how the posterior mean, $\bm{\bar{W}_t}$, evolves over training, via its (expected) spectrum. This algorithm learns all singular values concurrently, much like the pattern observed in the Meta-Learner.

\subsection{The choice of optimiser}
\label{subsection:linear-regression:optimiser}

How dependent are these results on the underlying optimisation algorithm in use? The experiments presented so far were all performed using an SGD optimiser, which the Learner deployed directly in service of the regression task at hand, and which the Meta-Learner deployed during outer training.

\begin{figure*}[th]
\begin{centering}
  \includegraphics[width=18cm]{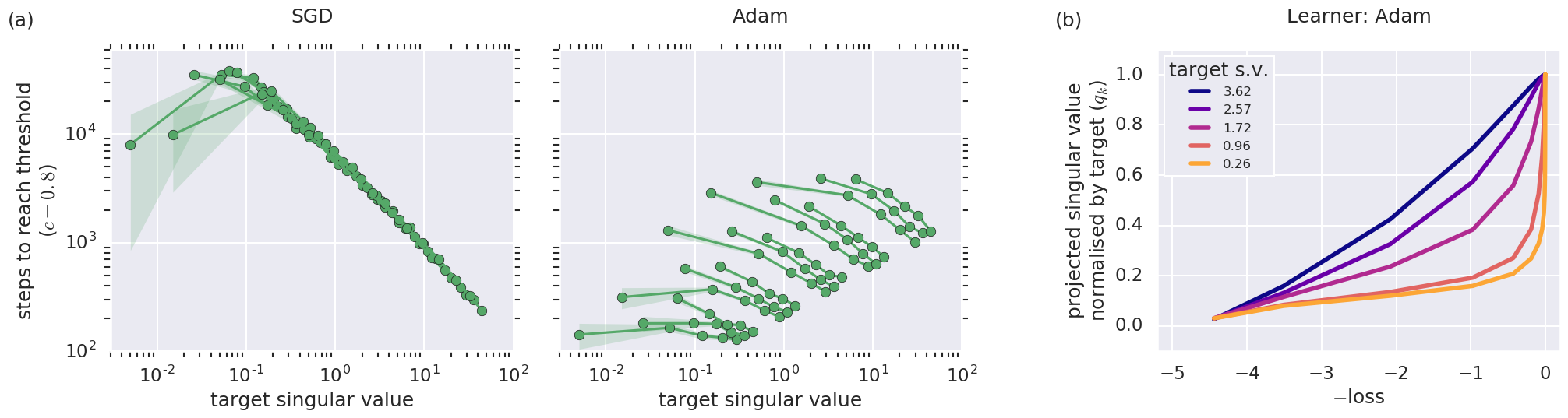}
	\caption{
		Learning dynamics of the Learner on 5D regression problems, when using SGD and (left) Adam optimisers (right).
		\textbf{(a)} Training steps to reach the 80\% threshold of each singular value (i.e., for $q_k(t) > 0.8$), as in Fig~\ref{fig:linear:learner:individual-modes}c. Each line shows the dynamics on target matrices with a particular spectrum. Both optimisers induce the same learning dynamics on each task: the dominant singular mode is learned first.
		\textbf{(b)} Effective spectrum of a 5D regression problem for the Adam-based Learner, over the course of training, as in Fig~\ref{fig:linear:spectrum-vs-performance}. Note that the use of Adam maintains the staggered pattern of learning singular modes.
		\label{fig:linear:learner:adam}
		}
\par\end{centering}
\end{figure*}

We repeated the same experiments using the Adam optimiser \cite{kingma2014adam}. While this changed the Learner's speed of learning \textit{across} regression problems---such that tasks with a larger norm of the matrix $\bm{W}$ were slower to learn---it did not change the general pattern of learning a \textit{given} regression problem: the dominant singular mode was always learned first, and smaller singular modes later (Fig~\ref{fig:linear:learner:adam}).

We observed no qualitative effects of using Adam rather than SGD on the behaviour of the Meta-Learner.

\subsection{Summary}

In this section, we compared how sample-inefficient deep Learners, and sample-efficient LSTM Meta-Learners (pre-configured through meta-learning) progressively capture the structure of a linear regression task. While Learners latch on to the singular modes progressively, in order of their singular values, the sample-efficient LSTM Meta-Learners estimate all the singular modes concurrently. These latter learning dynamics are congruent with the dynamics of Bayes-optimal linear regression algorithms.

\section{Experiment 2: Nonlinear regression}
\label{section:nonlinear-regression}

In our second experiment, we study a phenomenon reported by \citet{rahaman2018spectral} and \citet{xu2018training}. This work demonstrates that Deep ReLU and sigmoid networks discover structure in \textit{nonlinear} regression problems in a similarly staggered fashion: they uncover the low Fourier frequencies in the target function before the higher frequencies.

In what follows, we first replicate empirical results for a deep Learner for completeness (Section~\ref{subsection:nonlinear-regression:learner}), again setting up the standard experimental parameters and protocols. As in the previous Experiment, we then consider the dynamics by which a sample-efficient Meta-Learner---outer-trained to learn nonlinear regression problems---acquires structure while learning problems of the same form (Section~\ref{subsection:nonlinear-regression:meta-learner}). Once again, we save analysis of the learning process by which the Meta-Learner itself is constructed to Section~\ref{section:outer-loops}.

\subsection{Learner}
\label{subsection:nonlinear-regression:learner}

\subsubsection{Setup}

We consider nonlinear regression problems of the form $y = g(x) + \varepsilon$, where $\varepsilon \sim \mathcal{N}(0, \sigma_\varepsilon^2)$. We consider here only univariate functions $g: \mathbb{R} \rightarrow \mathbb{R}$ defined over the interval $[-0.5, 0.5]$, though the theory and experiments in previous work \cite{rahaman2018spectral, xu2018training, xu2018understanding} extend to the more general multivariate case (with larger support).

We train Learners on a single task, defined by sampling a scalar function $g$ to have a (boxcar) low-pass Fourier spectrum (with a zero DC term), defined using the first $K = 5$ modes of the Fourier series:
\begin{equation}
g(x) = \sum_{k=1}^{5} \sin(2 \pi k x + \phi_k)
\label{eqn:nonlinear-regression:g}
\end{equation}
This amounts to sampling the phases $\phi_k \sim \mathcal{U}[0, 2\pi)$. This function is then held fixed for the duration of training. All data is procedurally-generated, with $x \sim \mathcal{U}[-0.5, 0.5)$.

Details of the Learners can be found in Appendix~\ref{appendix:architectures:nonlinear:learner}. The objective is to minimise the $l_2$ loss between the network's prediction $\hat{y}$ and the target $y$.

\subsubsection{How to assess learning dynamics}
\label{subsubsection:nonlinear-regression:learner:how}

At each training step, $t$, we estimate the effective function represented by the Learner by feeding in a probe minibatch, $\bm{x_\mathrm{probe}}$, comprising $N = 40$ equispaced $x$ values over the $[-0.5, 0.5)$ interval\footnote{To ensure periodicity, we sample $N = 41$ equispaced points over $[-0.5, 0.5]$, drop the final one, and centre the values to have zero mean.}. From the network's set of outputs, $\bm{\hat{y_t}}$, we compute the Fourier coefficients via an FFT. 

Our interest is in the relationship between the complex Fourier coefficients of the ground-truth function, $\tilde g_k$, and those estimated from $\bm{\hat{y_t}}$, $\hat{\tilde g}_{k,t}$, where $k$ indexes the coefficient. Analogous to the projection onto singular modes computed in the first experiment, we compute the (normalised) projection of the estimated Fourier modes onto the true ones:
\begin{equation}
    q_k(t) = \frac
    {\mathrm{Re}
        \left(
            \langle \hat{\tilde{g}}_{k, t} , \tilde{g}_k \rangle
        \right)}
    {|\tilde{g}_k|^2}
\label{eqn:nonlinear-regression:projection}
\end{equation}
where $\langle f, g\rangle = f g^\ast$ is the complex inner product. This value of $q_k(t)$ approaches one when both the amplitude and the phase of the learned Fourier modes approach those of the true Fourier modes\footnote{We obtained similar results when we ignored the phase, and simply computed the ratio of the spectral energies for each $k$.}.

\subsubsection{Results}

\begin{figure*}[th]
\begin{centering}
	\includegraphics[width=18cm]{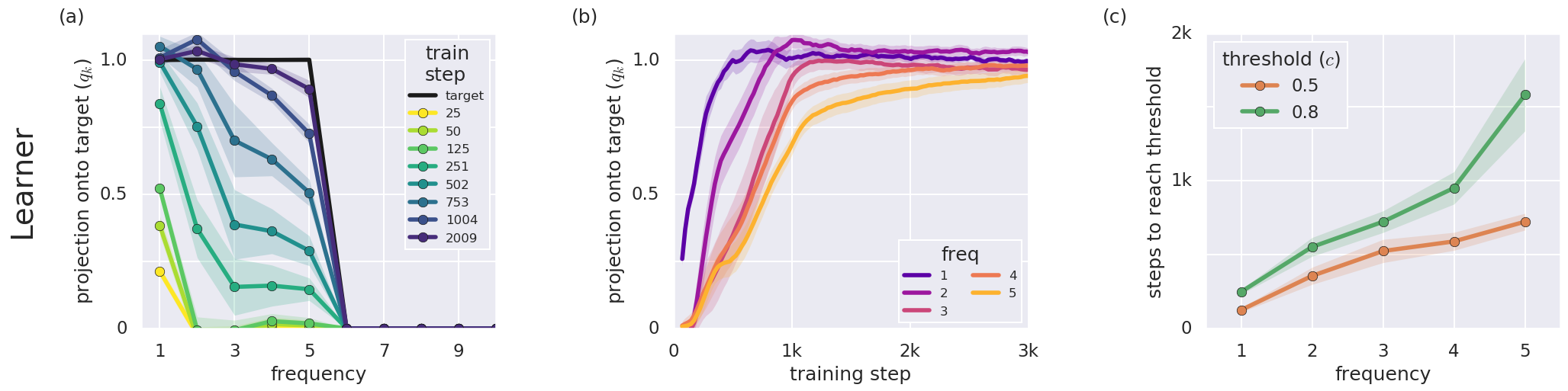}
  \caption{
		\textbf{(a)} Projection of the Learner's learned function onto the Fourier modes of the ground-truth function, $q_k(t)$, at different training times $t$. The low frequencies can be seen to converge to their targets earlier in training.
		\textbf{(b)} Learning dynamics of individual frequencies over training.
		\textbf{(c)} Training steps required for the Learner to reach a threshold of performance where $q_k(t) > c$, for different cutoff values $c$. This demonstrates that the learning speed of Fourier modes is slower for higher frequencies.
  \label{fig:nonlinear:learner:individual-freqs}
				}
\par\end{centering}
\end{figure*}

Figs~\ref{fig:nonlinear:learner:individual-freqs}a-b show the Learner's learning dynamics for individual frequencies. As per \citet{rahaman2018spectral} and \citet{xu2018training}, lower-frequency Fourier modes of $g$ (darker colours in Fig~\ref{fig:nonlinear:learner:individual-freqs}b) are learned faster, i.e.\ with learning curves that converge quicker. Conversely, higher-frequency Fourier modes of $g$ (lighter colours in Fig~\ref{fig:nonlinear:learner:individual-freqs}b) are learned slower. As in the previous section, we quantify the learning rate of these Fourier modes by measuring the number of training steps it takes for the learned modes to reach a threshold proportion of the target signal, $q_k(t) > c$, as in Fig~\ref{fig:linear:learner:individual-modes}c. Here it can be seen that the higher the frequency of the Fourier mode, the more steps it takes.

\subsection{Meta-Learner}
\label{subsection:nonlinear-regression:meta-learner}

\subsubsection{Setup}
\label{subsubsection:nonlinear-regression:meta-learner:setup}

We sought to compare these learning dynamics with those of a sample-efficient meta-learner, which had been explicitly configured to be able to learn new nonlinear regression problems.

To do this, we repeated the same experiment as in the previous Section, but now with LSTM Meta-Learners outer-trained on nonlinear regression problems (architecture details in Appendix~\ref{appendix:architectures:nonlinear:meta-learner}).

On each length-$T$ episode $i$ during outer-training, we presented the Meta-Learner with a new nonlinear regression problem, by sampling a new target function $g^{(i)}$. Each function had the same target Fourier spectrum
\footnote{We also ran experiments where the amplitudes of each Fourier mode were varied across episodes, i.e.\ where $g^{(i)}(x) = \sum_{k=1}^{K} a^{(i)}_k \sin(2 \pi k x + \phi_k)$, with each $a^{(i)}_k \sim \mathcal{U}[0, 1]$. Results were qualitatively identical for these experiments.}
, but different phases, $\phi_k^{(i)}$. We again sampled a sequence of inputs, $\lbrace x^{(i)}_t \rbrace_{t=1}^{T}$, where each $x^{(i)}_t \sim \mathcal{U}[-0.5, 0.5)$ for each time step $t \in \lbrace 1, ..., T \rbrace$. In turn, we computed the corresponding targets, $y^{(i)}_t = g^{(i)} ( x^{(i)}_t ) + \varepsilon^{(i)}_t$. The remainder of the setup was otherwise as described in Section~\ref{subsubsection:linear-regression:meta-learner:setup}. We used episode lengths of $T=40$.

\subsubsection{How to assess learning dynamics}
\label{subsubsection:nonlinear-regression:meta-learner:how}

We consider the inner dynamics here, by which the configured LSTM learns a solution to a new nonlinear regression problem. We leave analysis of outer-loop behaviour to Section~\ref{section:outer-loops}.

Our goal is to estimate the effective function realised by the LSTM at iteration $t$ of an episode $i$, which we denote $\hat{g}^{(i)}_t: \mathbb{R} \rightarrow \mathbb{R}$, and determine how this function changes with more observations (i.e., as $t$ increases). In particular, our interest is in the Fourier coefficients of $\hat{g}^{(i)}_t(\cdot)$, and how these compare to the Fourier coefficients of the target function $g^{(i)}_t(\cdot)$.

Our procedure for doing this is analogous to that described for the linear regression Meta-Learner, described in Section~\ref{subsubsection:linear-regression:meta-learner:how}. For this nonlinear case, the probe set of inputs at each time point $t$ was the equispaced minibatch $\bm{x_\mathrm{probe}}$ as described in Section~\ref{subsubsection:nonlinear-regression:learner:how}; we then apply an FFT to the respective outputs, and compute the normalised projection as in Equation~\eqref{eqn:nonlinear-regression:projection}. For each time, $t$, we average the projections over 2000 episodes (each with unique phases), and report statistics over 5 independently-trained instances of Meta-Learner LSTMs.

\subsubsection{Results: in-distribution}

\begin{figure*}[th]
\begin{centering}
	\includegraphics[width=18cm]{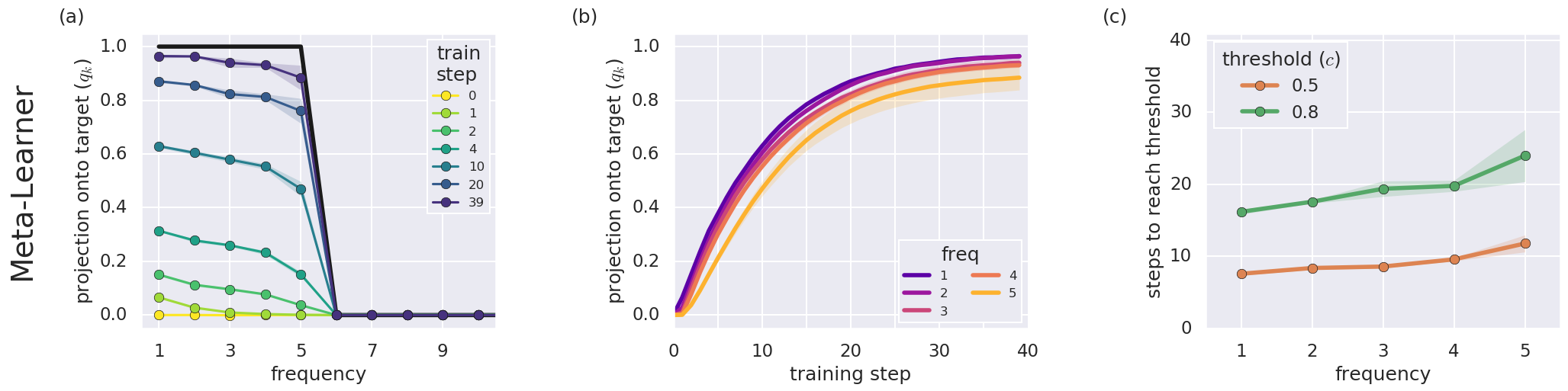}
  \caption{
	Learning dynamics of the Meta-Learner for different Fourier modes, as in Fig~\ref{fig:nonlinear:meta-learner:individual-freqs}.
  \label{fig:nonlinear:meta-learner:individual-freqs}
  }
\par\end{centering}
\end{figure*}

Fig~\ref{fig:nonlinear:meta-learner:individual-freqs} shows the Meta-Learner's learning dynamics for individual Fourier modes, in the same form as Fig~\ref{fig:nonlinear:learner:individual-freqs}. We notice again the same discrepancy described in the linear regression case: the Meta-Learner does \textit{not} show the Learner's pattern of learning Fourier modes with lower frequencies faster. Rather, the Meta-Learner learns all (in-distribution) frequencies almost simultaneously, with only a gentle increase in learning time for higher frequencies.

Putting these changes into perspective, the $K=5^\mathrm{th}$ frequency takes the Learner $6.4\times$ as long to learn than the $K=1$ term, while it causes the Meta-Learner to take only $1.5\times$ as long.
\footnote{Indeed, if one accounts for the incomplete learning of the higher frequencies at this point during outer-training, this factor drops to $1.2\times$ (Fig~\ref{appendix:fig:nonlinear:meta-learner:individual-freqs-normalised}).}

\begin{figure*}[th]
\begin{centering}
	\includegraphics[width=13cm]{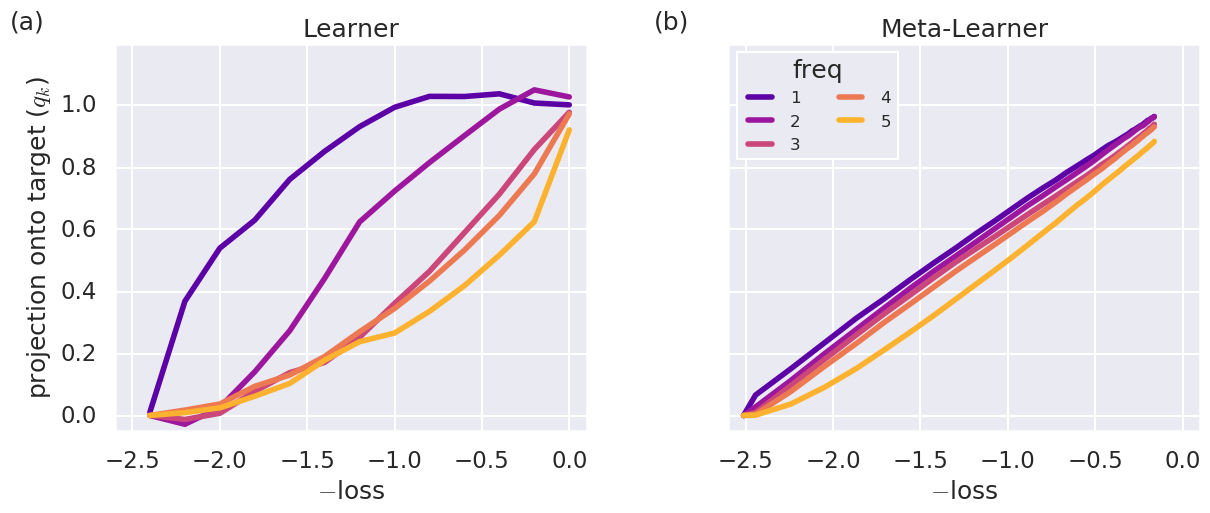}
	\caption{
		Effective portion of a nonlinear function learned over the course of training, broken down by Fourier components, for \textbf{(a)} the Learner, and \textbf{(b)} the Meta-Learner, as in Fig~\ref{fig:linear:spectrum-vs-performance}.
	\label{fig:nonlinear:spectrum-vs-performance}
	}
\par\end{centering}
\end{figure*}

As in the linear regression case, it is instructive to view the difference between the structure that the Learner and Meta-Learner exploit at different milestones of learning. We show this in Fig~\ref{fig:nonlinear:spectrum-vs-performance}, where it can be clearly seen how the Learner acquires the Fourier structure of nonlinear functions with a staggered time course, while the Meta-Learner acquires all the Fourier structure roughly concurrently.

As in the case of linear regression, this overall difference could not be attributed to differences in parameterisation between the Learner and the Meta-Learner (Fig~\ref{appendix:fig:nonlinear:learner:lstm:individual_freqs}).

\subsubsection{Results: out-of-distribution}
\label{section:nonlinear-regression:meta-learner:ood}

Finally, we observe that since the Meta-Learner is configured to have an effective prior over Fourier spectra, it suppresses structure outside the support of this distribution. 

We outer-trained a Meta-Learner LSTM on bandpass functions, where $a_3 = a_4 = a_5 = 1$, and $a_k = 0 \; \forall k$ otherwise. This Meta-Learner LSTM shows the same learning dynamics on the pass-band as described above, but all stop-band signals are not fit at all (Fig~\ref{appendix:fig:nonlinear:meta-learner:bandpass:individual-freqs}). This effect thus appears to be far more dramatic than the bias introduced for linear regression problems, which showed a gradual degradation of learning outside of the training distribution. 

\subsubsection{Relationship to Bayes-optimal inference}
\label{subsubsection:nonlinear-regression:bayes-optimal}

As in Section~\ref{subsubsection:linear-regression:bayes-optimal}, we again demonstrate that the Meta-Learner shows the same qualitative learning dynamics as can be expected from Bayes-optimal inference on the same distribution of tasks.

\begin{figure*}[th]
\begin{centering}
  \includegraphics[width=18cm]{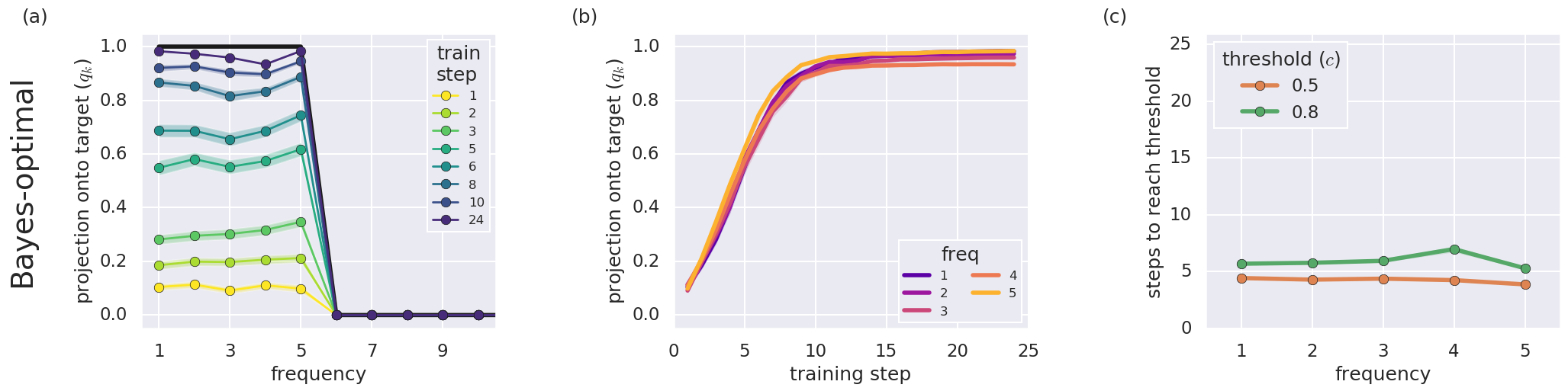}
	\caption{
	  Learning dynamics of Bayesian nonlinear regression for functions $g$ of the form given in Equation~\eqref{eqn:nonlinear-regression:g}. Results shown as for Fig~\ref{fig:nonlinear:meta-learner:individual-freqs}.
		\label{fig:nonlinear:optimal}
		}
\par\end{centering}
\end{figure*}

For a simple approximation of Bayes-optimal inference over functions $g$ of the form given in Equation~\eqref{eqn:nonlinear-regression:g}, we implemented a tabular look-up table of functions defined in phase space, i.e. where $\bm{\phi} \in [0, 2\pi]^K$. We subdivided this space into 16 bins along each axis, producing $16^{5} \approx 10^6$ functions. We use this to compute a posterior mean function, $\bar{g}_t$, after observing $t$ input-output pairs. In Fig~\ref{fig:nonlinear:optimal}, we show the expected projection of the posterior mean function onto the Fourier modes of the ground-truth function. This algorithm learns all Fourier modes concurrently, much like the pattern observed with the Meta-Learner.

\subsection{Summary}

In this section, we compared how sample-inefficient deep learners, and sample-efficient LSTM Meta-Learners (pre-configured through meta-learning) progressively capture the structure of a nonlinear regression task. While Learners latch on to the Fourier modes in order of their frequency---first the lower frequencies, then the higher ones---the sample-efficient LSTM Meta-Learners estimate all the Fourier modes concurrently. These latter dynamics are congruent with the dynamics of Bayes-optimal inference on the family of nonlinear regression problems on which the Meta-Learners have been trained.

\section{Experiment 3: Contextual bandits}
\label{section:contextual-bandits}

In our third experiment, we turn to the dynamics of reinforcement learning. \citet{schaul2019ray} describes a phenomenon that occurs during on-policy reinforcement learning: when function approximators are trained on tasks containing multiple contexts, there is an interference effect wherein improved performance in one context suppresses learning for other contexts. This occurs because the data distribution is under the control of the learner; as a policy for one context is learned, this controls behaviour in other contexts, and prevents the required exploration. This interference does not occur when one learns supervised, or off-policy, or when one trains separate systems for the separate tasks.

Here we first replicate the contextual bandit experiments in \citet{schaul2019ray} using a simple Learner (Section~\ref{subsection:bandits:learner}). We then consider the dynamics by which a sample-efficient Meta-Learner---outer-trained to learn contextual bandit problems---acquires structure while learning problems of the same form (Section~\ref{subsection:bandits:meta-learner}). In each case, we compare these learning dynamics to the learning behaviour of decoupled systems, for which the proposed interference should not occur. Analysis of the learning process by which the Meta-Learner itself is constructed is saved for Section~\ref{section:outer-loops}.

\subsection{Learner}
\label{subsection:bandits:learner}

\subsubsection{Setup}

We consider contextual bandit problems of the following form. We assume that there are $K_c$ contexts (whose identity is fully observed), and $K_a$ actions that are all available in all contexts. Within each context, there is a correct action, which, if taken, delivers a unit reward with probability $p_\mathrm{correct}$; if an incorrect action is taken, a unit reward is delivered with probability $p_{\mathrm{incorrect}}$. On each step, the context is sampled randomly.

For our experiments here, we use $K_c = K_a = 5$ contexts and actions. We also use non-deterministic rewards with $p_{\mathrm{correct}}=0.8$ and $p_{\mathrm{incorrect}}=0.2$ (though qualitatively identical results were obtained with values of $1$ and $0$ respectively, as used in \citet{schaul2019ray}). We explicitly only consider tasks where the contexts are in conflict with each other, i.e.\ where each context has a unique correct action.

We use a linear function approximator as our Learner: a simple network that takes the context $c$ as input (represented by a one-hot, $\bm{c}$), and outputs a policy given by
\begin{equation}
	\pi(\bm{a}|c) = \mathrm{softmax}(\bm{W}\bm{c} + \bm{b})
  \label{eqn:bandits:linear:policy}
\end{equation}
As described in \citet{schaul2019ray}, the biases $\bm{b}$ ``couple'' the respective contexts' learning processes, causing interference. We train the network via REINFORCE \cite{williams1992simple} and Adam (see Appendix~\ref{appendix:architectures:bandits:learner} for details; similar results obtain using SGD). 

We also compare the Learner to one where $K_c$ independent networks handle the policy for each of the $K_c$ contexts. For each network, we use a similar linear + softmax parameterisation as Eqn~\ref{eqn:bandits:linear:policy}; the $c^\mathrm{th}$ network is thus effectively tabular, as is the overall policy learned by this system.

\subsubsection{How to assess learning dynamics}

At each training step, $t$, we tracked the context-dependent policy $\pi_t(\cdot|c)$, and extracted the probability of taking the correct action in each context, $q_c(t) = \pi_t(a_{\mathrm{correct}}(c)|c)$. This gives the expected reward in context $c$ at time $t$ to be:
\begin{equation}
  \mathbb{E}[r_c(t)] =
				q_c(t) p_\mathrm{correct}
				+ (1 - q_c(t)) p_\mathrm{incorrect}
\label{eqn:bandits:expected-reward}
\end{equation}
where the (unsubscripted) expectation, $\mathbb{E}$, is over the stochasticity in the policy and the environment. This, in turn, gives the overall expected reward at time $t$ as:
\begin{equation}
				\mathbb{E}[r(t)] = \frac 1 {K_c} \sum_c \mathbb{E}[r_c(t)]
\end{equation}

As our interest was in relating the dynamics of the first-learned context to that of the second-learned context, and the third-, etc, we rank-ordered the contexts by their expected return over the course of training, $R_c$:
\begin{equation}
				R_c = \mathbb{E}_t \mathbb{E}[r_c(t)]
				\label{eqn:bandits:expected-return}
\end{equation}
Given the linearity of Eqn~\eqref{eqn:bandits:expected-reward} (and that $p_\mathrm{correct} > p_\mathrm{incorrect}$), this is equivalent to rank-ordering by the time-average of $q_c(t)$. With this ordering, we switch notation from the $c^\mathrm{th}$ context to the $k^\mathrm{th}$-learned context. Thus, for example, $q_1(t)$ refers to the probability of choosing the correct action at time $t$ for the context that happens to have been learned first.

\subsubsection{Results}

\begin{figure*}[th]
\begin{centering}
	\includegraphics[width=18cm]{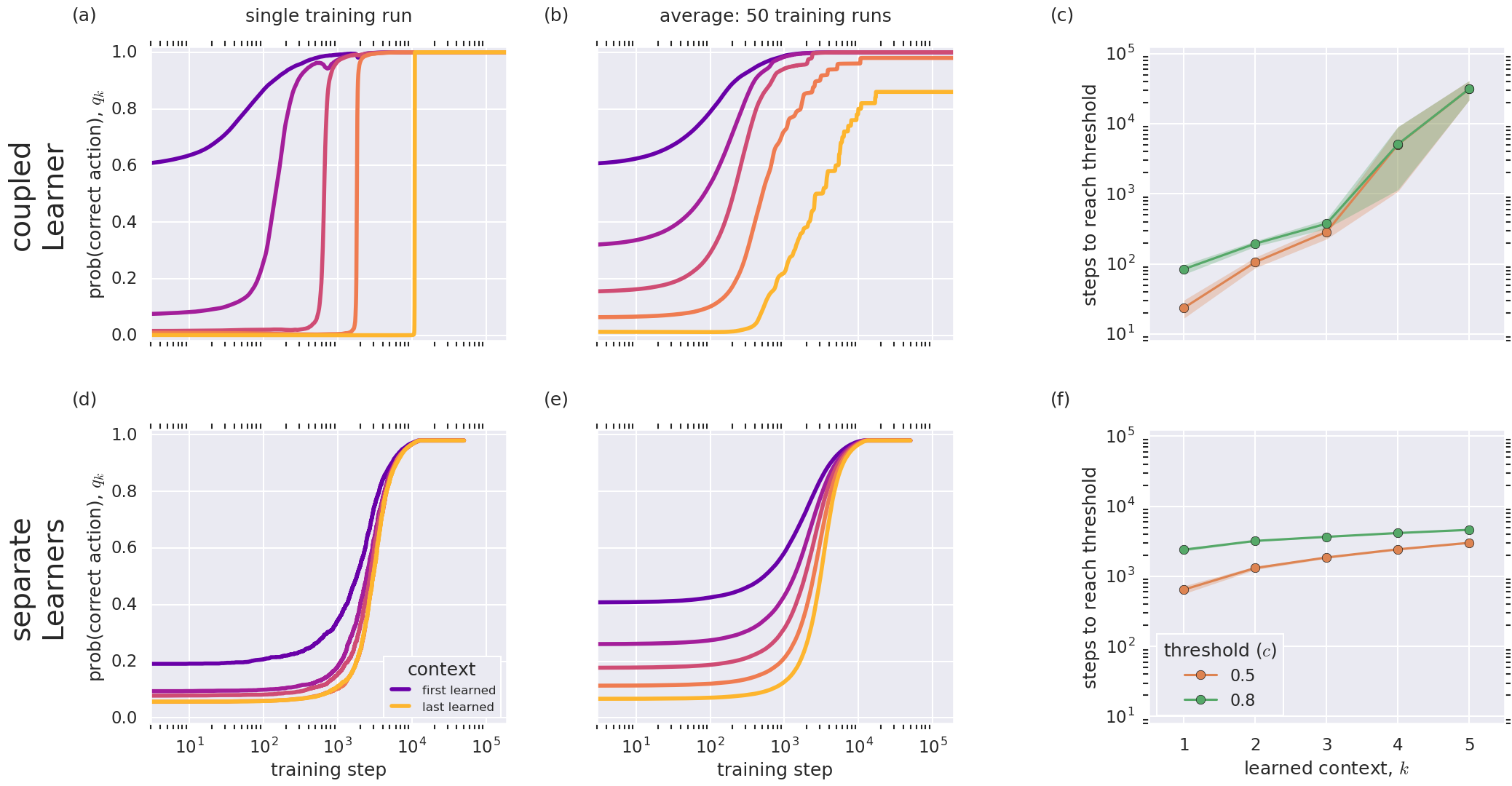}
	\caption{
	\textbf{(a)} Learning dynamics ($q_k(t)$) of the coupled Learner on a single training run. Contexts are ordered by the order in which they are learned. Note the logarithmic time scale on the abscissa.
	\textbf{(b)} As in (a), but averaged over 50 training runs. Data are averaged by the ordered indices, $k$, rather than the nominal indices, $c$.
	\textbf{(c)} Training steps required for the Learner to reach a threshold of performance where $q_k(t) > c$, for different cutoff values $c$. This demonstrates that later-learned contexts take exponentially longer to learn.
	\textbf{(d-f)} As in (a-c), except where separate Learners are used for each context.
	\label{fig:bandits:learner:individual-contexts}
	}
\par\end{centering}
\end{figure*}

Fig~\ref{fig:bandits:learner:individual-contexts} shows the Learner's learning dynamics for each successively learned context. When learning is coupled across tasks (as in the top row of Fig~\ref{fig:bandits:learner:individual-contexts}) the Learner takes exponentially longer, on average, to learn each successive context; the 5th context taking on average ${\sim}400\times$ as many steps to learn than the 1st context. When tasks are decoupled, so that learning occurs separately for each context (as in the bottom row of Fig~\ref{fig:bandits:learner:individual-contexts}), the delay between learning each successive context is far more gradual; the 5th context taking on average ${\sim}2\times$ as many steps to learn as the 1st context.

\subsection{Meta-Learner}
\label{subsection:bandits:meta-learner}

\subsubsection{Setup}
\label{subsubsection:bandits:meta-learner:setup}

Does this interference plague all of reinforcement learning? To test this, we sought to compare these learning dynamics with those of a sample-efficient meta-learner, which had been explicitly pre-trained to expect that policies for different contexts are independent of each other.

As in previous Sections, we built an LSTM Meta-Learner that, on every episode during outer-training, was presented with a new sample of a contextual bandits task. In particular, for each length-$T$ episode, $i$, we sampled a new mapping from contexts to correct actions. Importantly, for each episode, the correct action $a_{\mathrm{correct}}^{(i)}(c)$ for context $c$ was drawn \textit{independently} from the uniform distribution over actions. Thus an optimal learner should expect that data gleaned from actions taken in context $c$ are uninformative about a good policy in context $c^\prime$.

The remainder of the setup was as described in Sections~\ref{subsubsection:linear-regression:meta-learner:setup} and \ref{subsubsection:nonlinear-regression:meta-learner:setup}, with one notable change: rather than receiving the correct target for the previous time step, $y^{(i)}_{t-1}$, as an input at time $t$ (as is appropriate for supervised meta-learning), we instead feed the Meta-Learner the previous action taken, $a^{(i)}_{t-1}$ and the reward received, $r^{(i)}_{t-1}$, as per \citet{wang2016learning}. We outer-trained the Meta-Learner with REINFORCE and Adam. We used episode lengths of $T=100$, and $N_c = N_a = 5$ for consistency with the Learner. Further details are in Appendix~\ref{appendix:architectures:bandits:meta-learner}.

\subsubsection{How to assess learning dynamics}
\label{subsubsection:bandits:meta-learner:how}

We again consider the inner dynamics here, by which the configured LSTM learns a solution to a new contextual bandits problem. We leave analysis of outer-loop behaviour to Section~\ref{section:outer-loops}.

Our goal is to estimate the effective policy realised by the LSTM at iteration $t$ of an episode $i$, which we denote $\pi^{(i)}_t$, and determine how this policy changes with more observations (i.e., as $t$ increases). In particular, our interest is in the probabilities of taking the correct action in each context, $q_c^{(i)}(t) = \pi_t^{(i)}(a_{\mathrm{correct}}^{(i)}(c)|c)$. We did this in an analogous way to the previous Sections, by forking the LSTM for each time step $t$ during an episode, and probing the forked copies (counterfactually) with all possible contexts.

As with the Learner, we reordered contexts by their expected return over the course of training for that episode, $R^{(i)}_c$, defined analogously to Eqn~\eqref{eqn:bandits:expected-return}, and we use the index $k$ to refer to the $k^\mathrm{th}$-learned context in episode $i$.

We averaged results over 2000 episodes, and present statistics computed over 15 independently-trained Meta-Learners.


\subsubsection{Results}

\begin{figure*}[th]
\begin{centering}
	\includegraphics[width=18cm]{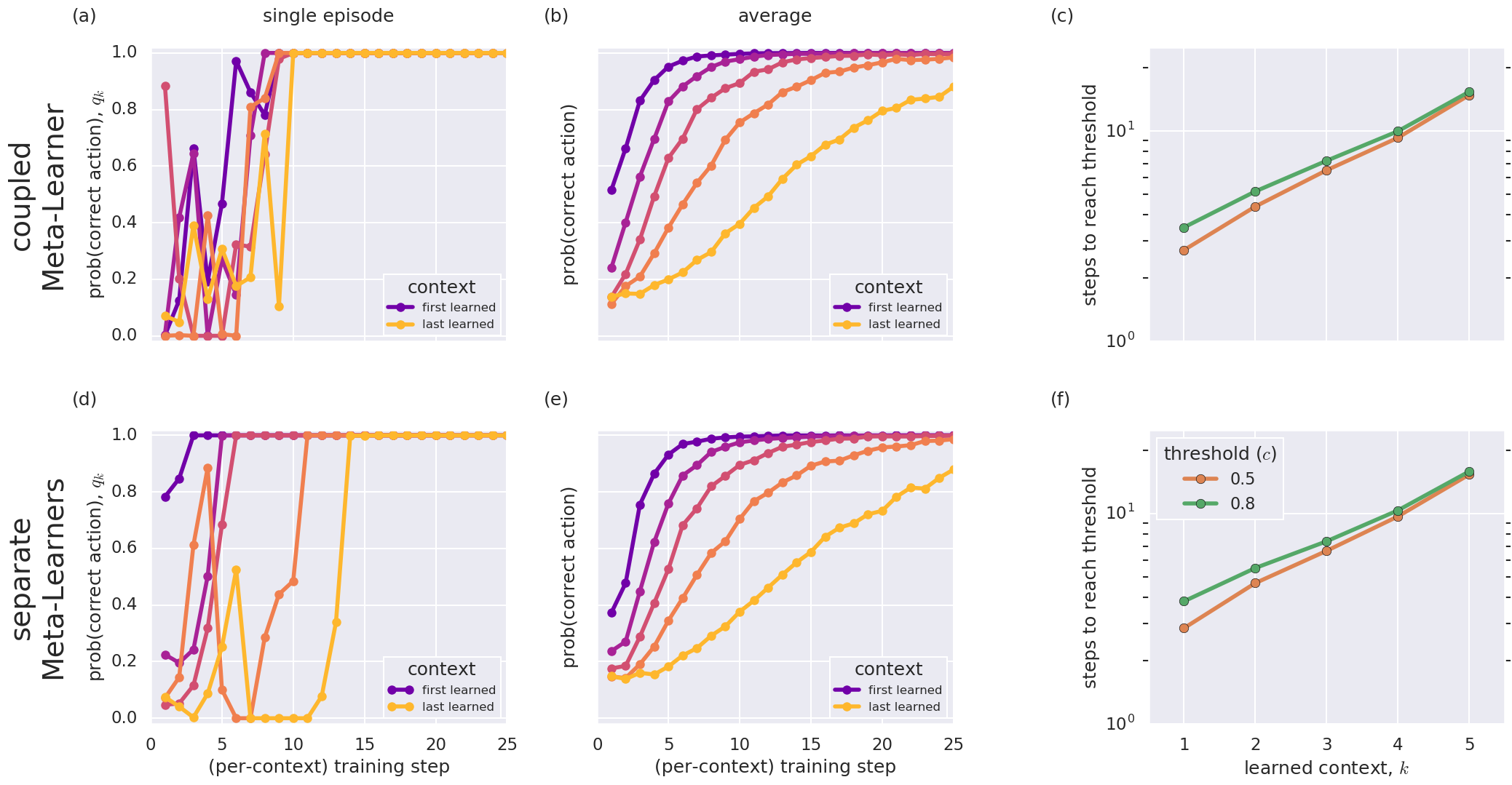}
  \caption{
	Learning dynamics of the Meta-Learner for successively-learned contexts, as in Fig~\ref{fig:bandits:learner:individual-contexts}.
  \label{fig:bandits:meta-learner:individual-contexts}
				}
\par\end{centering}
\end{figure*}

Fig~\ref{fig:bandits:meta-learner:individual-contexts} shows the Meta-Learner's learning dynamics for successively-learned contexts, in the same form as Fig~\ref{fig:bandits:learner:individual-contexts}. We notice again the same discrepancy described in the linear and nonlinear regression cases: while the Learner experiences a profound delay between learning successive contexts (Fig~\ref{fig:bandits:learner:individual-contexts}), this delay is relatively mild for the Meta-Learner (Fig~\ref{fig:bandits:meta-learner:individual-contexts}). Moreover, there is little difference between these dynamics and the case when the individual learning tasks are decoupled to be learned by separate Meta-Learners: in the coupled case, the 5th context takes on average $4.3\times$ as many steps to learn as the 1st context, compared with $3.9\times$ for the decoupled case. We note that a Bayes-optimal inference algorithm would be expected to show parity between the coupled and separate learning dynamics: as it would have prior knowledge that the contexts in the coupled case are indeed independent, inference over context-specific policies would factorise.

\subsection{Summary}

In this section, we compared how sample-inefficient reinforcement learners and sample-efficient LSTM Meta-Learners (pre-configured through meta-learning) progressively capture the structure of a contextual bandits task. While Learners experience interference between contexts---wherein learning each context-specific policy interferes with the learning of a policy for other contexts---this interference is not experienced by sample-efficient Meta-Learned LSTMs.

\section{Meta-Learners' Outer Loops}
\label{section:outer-loops}

The results we have presented so far showcase the behaviour of meta-learned LSTMs \textit{after} significant meta-training had taken place. This analysis was motivated by the desire to compare the learning trajectories taken by a family of sample-efficient learning algorithms with those taken by a family of sample-inefficient learning algorithms.

We now turn to the question: how do these sample-efficient learners themselves get constructed? What trajectories do they take through the space of learning algorithms?

\begin{figure*}[th]
\begin{centering}
	\includegraphics[width=18cm]{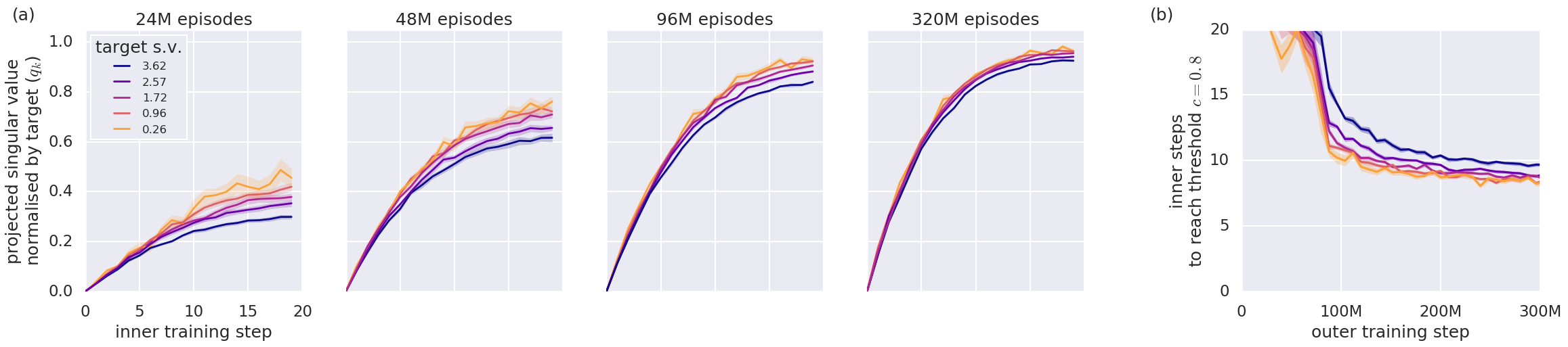}
  \caption{
	Outer learning dynamics of the Meta-Learner for the linear regression task.
	\textbf{(a)} Inner learning dynamics for linear regression problems at different stages during outer training. Curves shown for the single 5D linear regression task depicted in Fig~\ref{fig:linear:spectrum-vs-performance}. The final panel shows the qualitative behaviour described in Section~\ref{section:linear-regression}, whereby the meta-trained Meta-Learner learns all singular modes (roughly) concurrently.
	\textbf{(b)} Steps to reach threshold ($\mathrm{argmin}\, t$, s.t.\ $q_k(t) > 0.8$), as outer training progresses. Note that smaller singular values are learned faster during outer-training.
  \label{fig:linear:meta-learner:outer}
}
\par\end{centering}
\end{figure*}

\begin{figure*}[th]
\begin{centering}
	\includegraphics[width=18cm]{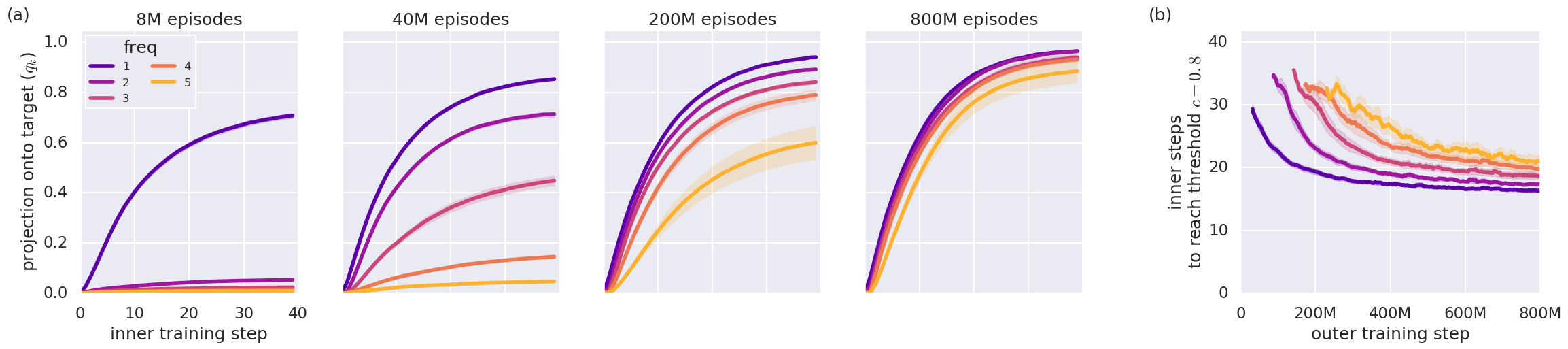}
  \caption{
	Outer learning dynamics of the Meta-Learner for the nonlinear regression task. Results presented in the same format as Fig~\ref{fig:linear:meta-learner:outer}. The final panel of (a) shows the qualitative behaviour described in Section~\ref{section:nonlinear-regression}, whereby the meta-trained Meta-Learner learns all Fourier modes (roughly) concurrently. Note the difference in time scale between the outer training of the coupled Meta-Learner and separate Meta-Learners, which shows the presence of interference early in meta-training.
  \label{fig:nonlinear:meta-learner:outer}
}
\par\end{centering}
\end{figure*}

\begin{figure*}[th]
\begin{centering}
	\includegraphics[width=18cm]{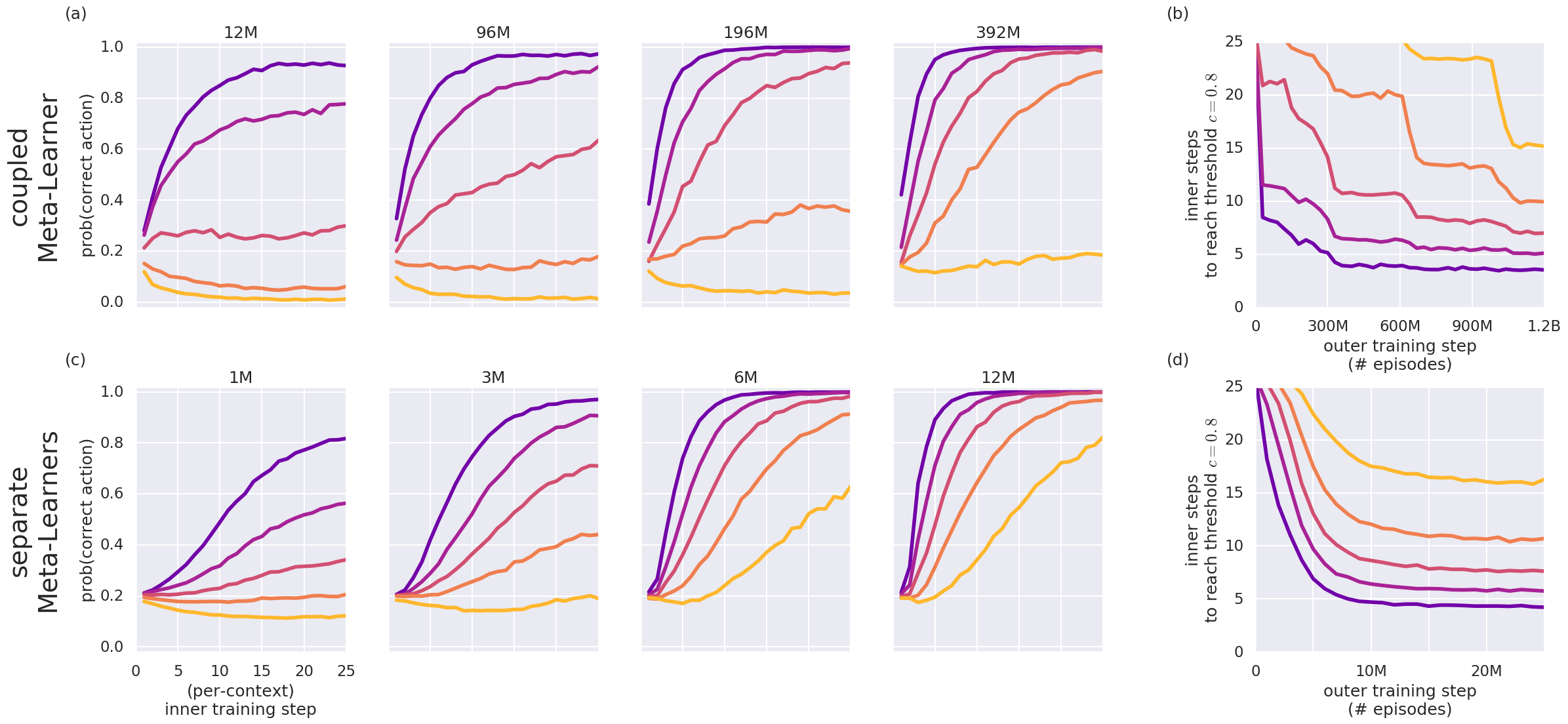}
  \caption{
	Outer learning dynamics of the Meta-Learner for the bandits regression task. Results presented in the same format as Fig~\ref{fig:linear:meta-learner:outer}. The rightmost panels of (a) and (c) shows the qualitative behaviour described in Setion~\ref{section:contextual-bandits}, whereby the meta-trained Meta-Learner learns all contexts without interference, i.e.\ in a near identical manner across coupled and separate conditions.
  \label{fig:bandits:meta-learner:outer}
}
\par\end{centering}
\end{figure*}

For each of the three experimental setups, we measured how the \textit{inner} learning trajectories progressed at different stages during \textit{outer} learning. These results are shown in Figs~\ref{fig:linear:meta-learner:outer}--\ref{fig:bandits:meta-learner:outer}.

Between these results, we note that the patterns of outer learning are not consistent across the three experiments. 

In the case of nonlinear regression and contextual bandits, the pattern of outer learning of the Meta-Learner is staggered, thus sharing a resemblance with the learning process of the Learner itself. For nonlinear regression, Fig~\ref{fig:nonlinear:meta-learner:outer} shows that early in meta-training, the inner learner is able to estimate the low-frequency structure within each task, but it takes many more steps of outer training until the inner learner is capable of estimating the high-frequency structure. This staggered meta-learning trajectory is analogous to the learning trajectory of the Learner, which acquires the Fourier structure in a similarly ordered fashion (albeit for a single task). This staggered structure is also observed for meta-training on the contextual bandit experiment: Fig~\ref{fig:bandits:meta-learner:outer} shows that there is significant interference between contexts early during meta-training, which is gradually overcome over iterations of the outer loop. 

In contrast to these phenomena, the outer learning of the Meta-Learner during linear regression follows an \textit{opposite} pattern. Fig~\ref{fig:linear:meta-learner:outer} shows that early in meta-training, the inner learner is better able to estimate the lesser singular modes than the greater ones. The ability of the Meta-Learner to estimate singular modes is thus staggered in the opposite direction to that of the Learner, which, early in training, has a better estimate of the greater singular modes than the lesser ones.

Notwithstanding these observations, the learning process of the Learner and the outer learning process of the Meta-Learner are not directly comparable. This is because the task being learned by these respective systems are fundamentally different. For example, in the linear regression problem, the Learner is being configured to solve a single regression problem, while the Meta-Learner, during meta-training, is being configured to be a generic\footnote{(at least, generic on the distribution of regression problems in the meta-training distribution)} linear regression algorithm. Solutions to these two tasks are radically different from one another; for example, we typically express the optimal solution to a single regression task as a matrix (viz., as the least squares \textit{solution}), while we typically express the optimal solution to generic linear regression via a highly non-linear formula (viz., as the least squares \textit{algorithm}, given in a reduced form in Eqns~\ref{eq:linear:bayes:1}--\ref{eq:linear:bayes:3}). As such, there may be no reason to expect any congruence between the Learners' learning dynamics and the Meta-Learners' outer learning dynamics, or indeed to expect similar outcomes across the three experiments, as the optimal solutions to the respective meta-tasks may look radically different.

\section{Discussion}
\label{section:discussion}

In this work we have shown three convergent pieces of evidence that meta-trained, sample-efficient Meta-Learners pursue systematically different learning trajectories from sample-inefficient Learners when applied to the same tasks.

Our goal here has not been to argue that the meta-learning systems are somehow better than the learners. As is well-known, (meta-trained) meta-learners' sample-efficiency comes at several costs, such as their high degree of task-specificity and the expense of meta-training. These are not issues which we attempt to address here. Rather, our goal has been to show that the process by which machine learners acquire a task's structure can differ dramatically once strong priors come into play.

We do not yet know the degree to which the discrepancies we identify between Learners and Meta-Learners extend to other task structures. We chose to focus on these three patterns---staggered learning of singular structure in linear regression, staggered learning of Fourier structure in nonlinear regression, and interference between context-specific policies in reinforcement learning---as these have already been described in the existing literature on deep learning and reinforcement learning. There is much to be discovered about how these patterns extend, for example, to the learning of hierarchical or latent structure, or whether they manifest in the learning of rich, complex real-world tasks.

While we have focused on memory-based Meta-Learners, there is additional work to be done to characterise the learning dynamics of other sample-efficient learning systems, including other meta-learning systems. Our expectation is that if these systems behave as if they are performing amortised Bayes-optimal inference with appropriately-calibrated priors, then the learning dynamics should resemble those of the LSTM Meta-Learners we have described here. On the other hand, the learning dynamics may diverge if these models have very strong priors, if they are being pushed to behave out-of-distribution, if they have pronounced capacity limitations, or if they are sufficiently general that their first stage of learning involves producing a complex inference about which task needs to be performed. All of these may hold, for instance, when considering human learners.

We believe that these findings are likely to have implications for curriculum learning. Whether one considers designing curricula for animals \cite{skinner1958reinforcement, peterson2004day}, humans \cite{beauchamp1968curriculum, hunkins2016curriculum, nation2009language}, or machines \cite{elman1993learning, sanger1994neural, krueger2009flexible, bengio2009curriculum, kumar2010self, lee2011learning, shrivastava2016training, weinshall2018theory}, the promise of this field is that one can accelerate learning by leveraging information about how the learner learns and how the task is constructed. By putting these together, one can decompose the task into chunks that are more easily learnable for the learner at hand. We have described how different machine learning systems can follow radically different learning trajectories depending on their priors; we may thus need to design radically different curricula for each to accelerate their respective learning.

These ideas may also connect to the challenges of how to bring humans into the loop of machine learning and machine behaviour. Direct human input has been proposed both as a means of accelerating machine learning \cite{ng1999policy, schaal1999imitation, abbeel2004apprenticeship, wilson2012bayesian, lin2017explore}, as well as a way of solving the value-alignment problem \cite{russell2015research, amodei2016concrete, hadfield2016cooperative, bostrom2017superintelligence, christiano2017deep, fisac2017pragmatic, leike2018scalable}. Either way, this research programme aims to enrich the pedagogical relationship between humans and machines. Successful pedagogy, however, depends on teachers having a rich model of their what their students know and how they learn \cite{shafto2014rational}\footnote{and also depends on students having a rich model of teachers' behaviour to infer their communicative intent.}. Through our work, we have shown that different machine learning systems pursue different learning trajectories; thus, if human teachers expect DL and RL systems to learn like sample-efficient human learners, their teaching strategies are likely to be miscalibrated. This may underlie recent observations made by \citet{ho2018people} that humans provide corrective feedback to virtual RL agents that leads them astray. The authors' interpretation was that humans' feedback is better described as a communication signal, rather than a reinforcement signal; in our framing, this comes down to a difference between the learning trajectories that different learners are expected to pursue. If we are committed to becoming better teachers, we will need to gain a better understanding of the specific learning trajectories that each of our machine pupils are apt to take. 

\section*{Acknowledgements}
Thanks to all the people who offered insightful comments on this work while preparing it, including Avraham Ruderman, Agnieszka Grabska-Barwinska, Alhussein Fawzi, Kevin Miller, Pedro Ortega, Jane Wang, Tom Schaul, and Matt Botvinick.

\appendix
\appendixpage
\renewcommand\thefigure{\thesection\arabic{figure}}    

\section{Network architectures}
\setcounter{figure}{0}   

\subsection{Linear regression: Learner}
\label{appendix:architectures:linear:learner}

For our Learner, we use 2-layer MLPs equipped with SGD; we use no nonlinearities, 10 hidden units, truncated normal initialisation ($\sigma = 0.1$), and learning rate $10^{-3}$. All data is procedurally-generated. We train with minibatch size 100. For the 2D problem, we train for 4k minibatches; for the 5D problem, we train for 16k minibatches.

Where we use Adam (Section~\ref{subsection:linear-regression:optimiser}), we use an identical setup, with a learning rate of $10^{-3}$.

\subsection{Linear regression: Meta-Learner}
\label{appendix:architectures:linear:meta-learner}

For our Meta-Learner, we use an LSTM equipped with SGD; we use 64 hidden units, and learning rate $10^{-2}$. We train with minibatch size 200. We use $T = 20$ steps per episode, and outer-trained over 80M training episodes. 

Where we use Adam (Section~\ref{subsection:linear-regression:optimiser}), we use a learning rate of $10^{-4}$, and outer-trained over 8M training episodes. 

\subsection{Nonlinear regression: Learner}
\label{appendix:architectures:nonlinear:learner}

For our Learner, we use 6-layer MLPs equipped with Adam; we use ReLU nonlinearities, 256 hidden units, and learning rate $10^{-4}$. We train with minibatch size 40, and train over 5k minibatches.

\subsection{Nonlinear regression: Meta-Learner}
\label{appendix:architectures:nonlinear:meta-learner}

For our Meta-Learner, we use an LSTM equipped with Adam; we use 64 hidden units, and learning rate $10^{-4}$. We train with minibatch size 200. We use $T = 40$ steps per episode, and outer-trained over 800M episodes. 

\subsection{Contextual bandits: Learner}
\label{appendix:architectures:bandits:learner}

For our Learner, we use a linear + softmax network. We initialise the weights with a truncated normal distribution ($\sigma = 1 / \sqrt{5}$), including a bias (to induce interference \cite{schaul2019ray}) and train using REINFORCE and the Adam optimiser with learning rate $10^{-4}$. We train with minibatch size 200, and train up to when the expected reward, $\mathbb{E}[r(t)]$, reaches 0.98. Similar results were obtained using the SGD optimiser in place of Adam. 

\subsection{Contextual bandits: Meta-Learner}
\label{appendix:architectures:bandits:meta-learner}

For our Meta-Learner, we use an LSTM equipped with REINFORCE and Adam; we use 64 hidden units, and learning rate $10^{-4}$. We train with minibatch size 200. We use $T = 125$ steps per episode, and outer-trained over 2B episodes. 



\begin{figure*}[th]
\begin{centering}
	\includegraphics[width=18cm]{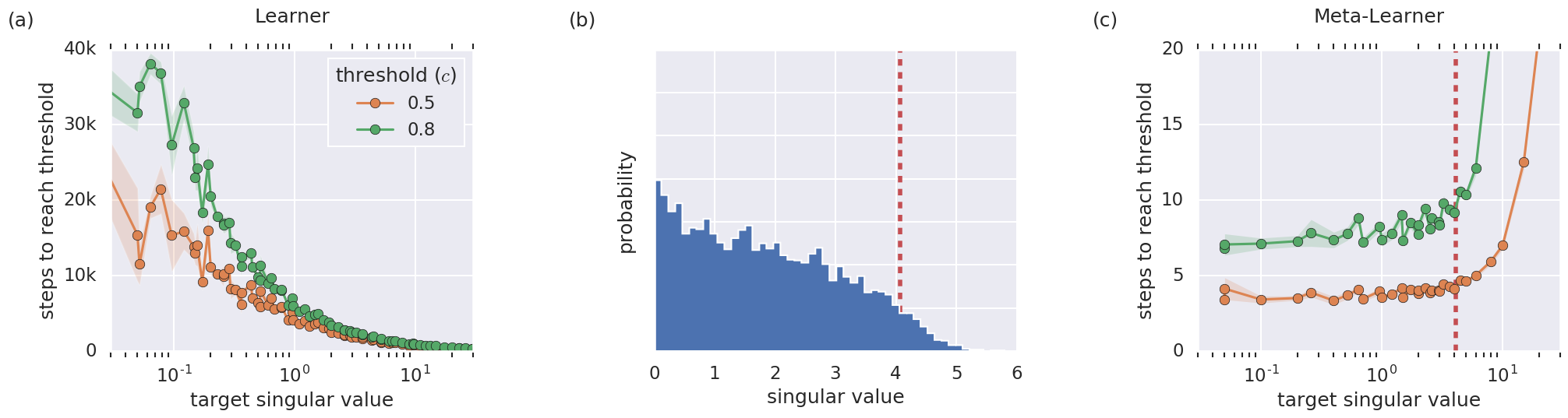}
	\caption{
		\textbf{(a)}
		Learning dynamics of the Learner for singular modes of different singular values, for 5D matrices (as in Fig~\ref{fig:linear:learner:individual-modes}c).
		\textbf{(b)} 
		Samples from marginal distribution of singular values for 5D matrices from the standard matrix normal distribution (as in Fig~\ref{fig:linear:distribution-of-spectra}b.)
		\textbf{(c)} 
		Learning dynamics of the Meta-Learner for singular modes of different singular values, for 5D matrices (as in Fig~\ref{fig:linear:meta-learner:individual-modes}c).
	\label{appendix:fig:linear:5D}
	}
\par\end{centering}
\end{figure*}

\begin{figure*}[th]
\begin{centering}
	\includegraphics[width=18cm]{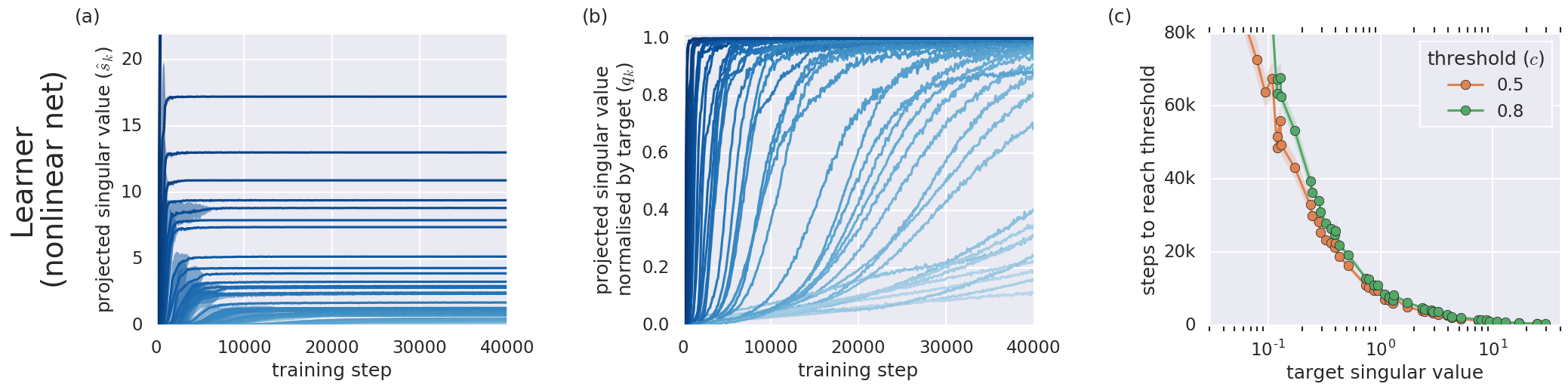}
	\caption{
		Learning dynamics of a Learner using a nonlinear network, rather than a linear one. We use here a 3-layer MLP, with 10 hidden units per layer, and ReLU nonlinearities (except for the last layer). Results shown for 2D linear regression problems, as in Fig~\ref{fig:linear:learner:individual-modes}.
	\label{appendix:fig:linear:nonlinear-net}
	}
\par\end{centering}
\end{figure*}

\begin{figure*}[th]
\begin{centering}
	\includegraphics[width=6cm]{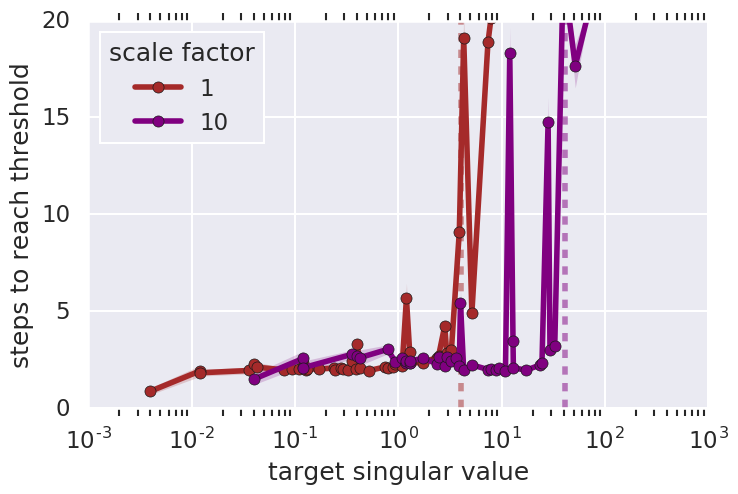}
	\caption{
	Meta-Learners outer-trained on scaled distributions of singular modes show the same qualitative (inner) learning dynamics, shifted accordingly. Here we show results for Meta-Learners trained with $p(\bm{W}) = \mathcal{MN}(\bm{0}, \alpha^2 \bm{I}, \alpha^2 \bm{I})$, where the scale factor $\alpha$ is either 1 or 10. Vertical lines show the $95^\mathrm{th}$ percentile of the distribution of singular values for each training distribution.
  \label{appendix:fig:linear:meta-learner:different-scales}
	}
\par\end{centering}
\end{figure*}

\begin{figure*}[th]
\begin{centering}
	\includegraphics[width=18cm]{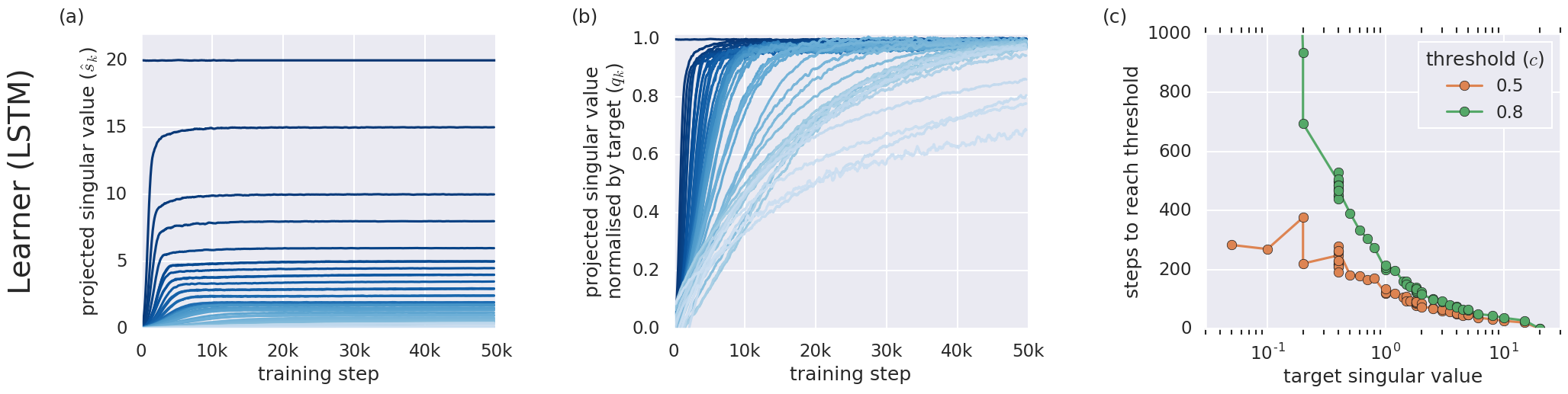}
	\caption{
	    Learning dynamics for a Learner parameterised as an LSTM on 2D linear regression tasks. Results presented as in Fig~\ref{fig:linear:learner:individual-modes}. The setup here was identical to the Meta-Learner (i.e., with precisely the same parameterisation), except instead of sampling a new target matrix $\bm{W^{(i)}}$ on each episode, the target matrix was kept fixed throughout the whole training process. We show here the behaviour of the learned function at the end of each ``episode'' (i.e., via the input-output function expressed at the final time step of the LSTM). Nevertheless, the results were qualitatively identical for all ``inner'' time steps. Despite the congruency between this parameterisation of the Learner, and that of the Meta-Learner, the learning dynamics remain different.
	\label{appendix:fig:linear:learner:lstm:individual_modes}
	}
\par\end{centering}
\end{figure*}


\begin{figure*}[th]
\begin{centering}
	\includegraphics[width=13cm]{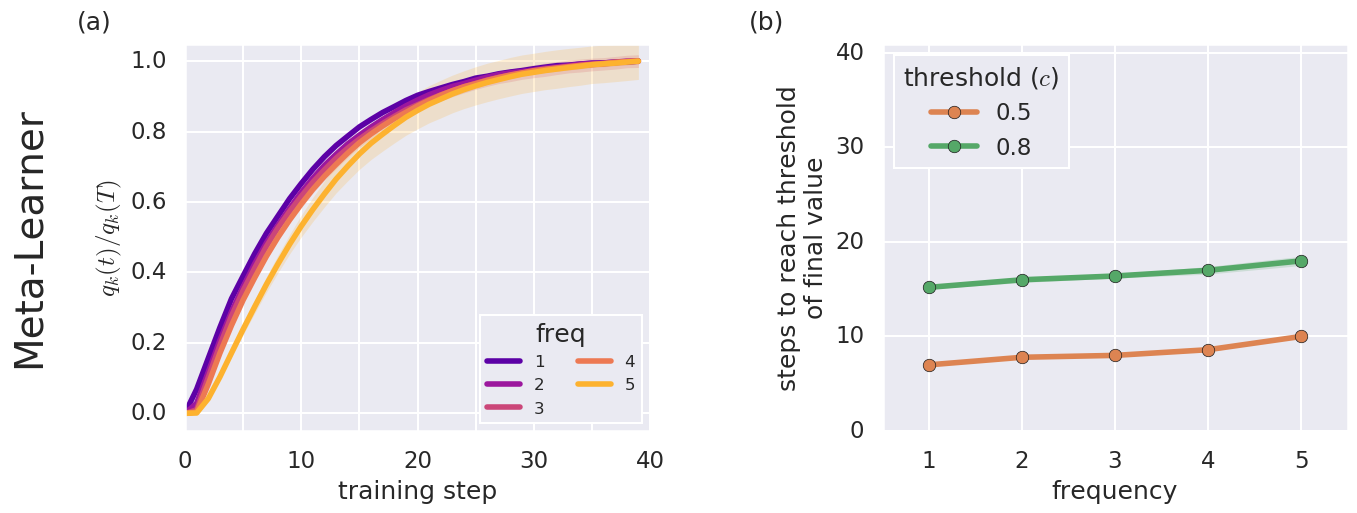}
  \caption{
			Learning dynamics of the Meta-Learner for different Fourier modes, as in Fig~\ref{fig:nonlinear:meta-learner:individual-freqs}. Since we terminated outer-training of the Meta-Learner before convergence (see Section~\ref{section:outer-loops}), inner learning of higher frequencies did not typically reach the target within the length of the episode. Here we adjust for this by measuring time constants of the curves shown in Fig~\ref{fig:nonlinear:meta-learner:individual-freqs}b with respect to their asymptotic value, not with respect to their target value.
  \label{appendix:fig:nonlinear:meta-learner:individual-freqs-normalised}
				}
\par\end{centering}
\end{figure*}

\begin{figure*}[th]
\begin{centering}
	\includegraphics[width=18cm]{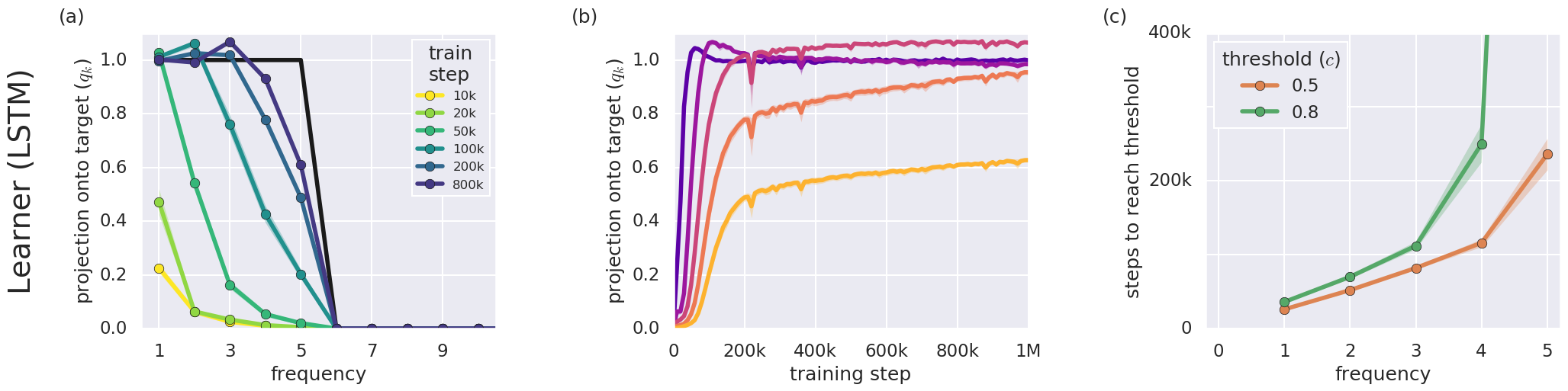}
	\caption{
	    Learning dynamics for a Learner parameterised as an LSTM on nonlinear regression tasks. Results presented as in Fig~\ref{fig:nonlinear:learner:individual-freqs}. The setup here mirrored that presented in the control experiment shown in Fig~\ref{appendix:fig:linear:learner:lstm:individual_modes}, i.e.\ using an LSTM as with the Meta-Learner, except instead of sampling a new function $g^{(i)}$ on each episode, the function was kept fixed throughout the whole training process. Despite the congruency between this parameterisation of the Learner, and that of the Meta-Learner, the learning dynamics remain different.
	\label{appendix:fig:nonlinear:learner:lstm:individual_freqs}
	}
\par\end{centering}
\end{figure*}

\begin{figure*}[th]
\begin{centering}
  \includegraphics[width=13cm]{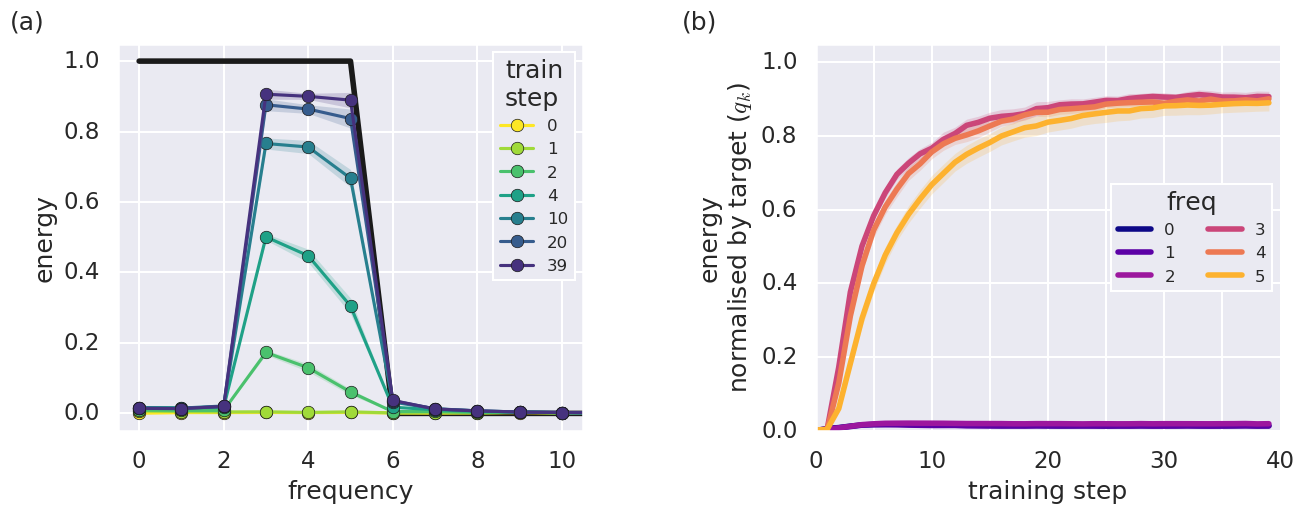}
	\caption{
	Learning dynamics of the Meta-Learner for different Fourier modes, when outer-trained to have a pass-band between $K=3$ and $K=6$. Results shown here are for nonlinear regression tasks where there is low-pass energy for $K < 3$, but the Meta-Learner does not learn this structure. Panels in the same format as in Fig~\ref{fig:linear:meta-learner:individual-modes}a-b.
	\label{appendix:fig:nonlinear:meta-learner:bandpass:individual-freqs}
	}
\par\end{centering}
\end{figure*}



\bibliographystyle{plainnat}
\bibliography{references}

\end{document}